\documentclass[journal]{IEEEtran}

\ifCLASSINFOpdf

\else

\fi
\usepackage{ulem} 
\usepackage{graphicx}
\usepackage{subfigure}
\usepackage{amsmath}
\usepackage{amssymb}
\usepackage{multirow}
\usepackage{algorithm}
\usepackage{algorithmicx}
\usepackage{algpseudocode}
\usepackage{verbatim}
\usepackage{bm}
\usepackage{makecell}
\usepackage{url}
\usepackage{multicol}
\usepackage{threeparttable}
\usepackage{xcolor}
\usepackage{soul}
\usepackage[justification=centering]{caption}

\makeatletter

\newcommand{\Rmnum}[1]{\expandafter\@slowromancap\romannumeral #1@}
\makeatother

\usepackage{color}
\usepackage{ulem}
\begin{document}    
\title{P$^3$-LOAM: PPP/LiDAR Loosely Coupled SLAM with Accurate 
Covariance Estimation and Robust RAIM in Urban Canyon Environment}

\author{Tao Li\thanks{Tao Li, Ling Pei, Yan Xiang, Qi Wu, Songpengcheng Xia, 
Lihao Tao, and Wenxian Yu are with 
Shanghai Key Laboratory of Navigation and Location-based Services, 
School of Electronic Information and Electrical Engineering, 
Shanghai Jiao Tong University, Shanghai, China, 200240 
(e-mail: firstname.lastname@sjtu.edu.cn). 
Corresponding author: Ling Pei (e-mail: ling.pei@sjtu.edu.cn)},
Ling Pei,
Yan Xiang,
Qi Wu,
Songpengcheng Xia,
Lihao Tao,
and Wenxian Yu}

\markboth{IEEE SENSORS JOURNAL}%
         {Shell \MakeLowercase{\textit{et al.}}: 
         Bare Demo of IEEEtran.cls for Journals}

\maketitle

\begin{abstract}
  Light Detection and Ranging (LiDAR) based 
  Simultaneous Localization and Mapping (SLAM) 
  has drawn increasing interests in autonomous driving.
  However, LiDAR-SLAM suffers from accumulating errors 
  which can be significantly mitigated 
  by Global Navigation Satellite System (GNSS).
  Precise Point Positioning (PPP), an accurate GNSS operation mode independent of 
  base stations, gains growing popularity in unmanned systems. 
  Considering the features of the two technologies, LiDAR-SLAM and PPP,
  this paper proposes a SLAM system, namely P$^3$-LOAM 
  (PPP based LiDAR Odometry and Mapping) 
  which couples LiDAR-SLAM and PPP.
  For better integration, we derive LiDAR-SLAM positioning covariance 
  by using Singular Value Decomposition (SVD) Jacobian model, 
  since SVD provides an explicit analytic solution of Iterative Closest Point (ICP), 
  which is a key issue in LiDAR-SLAM.
  A novel method is then proposed to evaluate the estimated LiDAR-SLAM covariance.

  In addition, to increase the reliability of GNSS in urban canyon environment, we develop 
  a LiDAR-SLAM assisted GNSS Receiver Autonomous Integrity Monitoring (RAIM) algorithm.
  Finally, we validate P$^3$-LOAM with UrbanNav, 
  a challenging public dataset in urban canyon environment.
  Comprehensive test results prove that, in terms of accuracy and availability, P$^3$-LOAM outperforms benchmarks such as 
  Single Point Positioning (SPP), PPP, LeGO-LOAM, SPP-LOAM, 
  and the loosely coupled navigation system proposed by the publisher of UrbanNav.
\end{abstract}


\IEEEpeerreviewmaketitle

\section{Introduction}
\IEEEPARstart{L}{ight} Detection and Ranging based Simultaneous Localization and Mapping
(LiDAR-SLAM) systems including 2D and 3D LiDAR-SLAM have been studied for decades. 
2D LiDAR-SLAM comprises 
FastSLAM \cite{montemerlo2002fastslam}, Hector-SLAM \cite{Kohlbrecher2011A}, 
GMapping \cite{Grisetti2007Improved}, 
KartoSLAM \cite{KartoSLAM}, LagoSLAM (Linear
Approximation for Graph-based SLAM) \cite{2011A} and so on, 
while 3D LiDAR-SLAM includes LOAM (LiDAR Odometry and Mapping) \cite{Ji2014LOAM}, 
LeGO-LOAM (Lightweight and Ground-Optimized LiDAR Odometry and Mapping) \cite{legoloam2018}, 
Suma \cite{Jens2018Efficient}, and Suma++ \cite{Xieyuanli2019SuMa}, among others.
These advanced systems aim to achieve self-motion estimation with low drift and high precision 
in the areas which require accurate positioning.
In general, though the cost of 3D LiDAR-SLAM is much higher, 
it is much more accurate and robust than 2D LiDAR-SLAM, 
especially in large-scale outdoor environment where 2D LiDAR-SLAM generally fails to work. 
Therefore, we focus on 3D LiDAR-SLAM in this paper.

Currently, the LiDAR-SLAM framework mainly involves two parts: frontend and backend. 
The frontend registers two frames of point cloud data and then obtains the transformation matrix.
The frontend primarily includes three kinds of algorithms:
Iterative Closest Point (ICP) \cite{besl1992method} 
as well as its extensions \cite{censi2008icp}, \cite{low2004linear}, 
Correlation Scan Matching (CSM) \cite{olson2009real}, 
and Normal Distributions Transform (NDT) \cite{magnusson2009three}.

As for the backend of LiDAR-SLAM, it serves to mitigate errors accumulated from the frontend.
Early studies utilized filtering methods 
such as Extended Kalman Filtering (EKF) \cite{anderson2012optimal}, 
Unscented Kalman Filtering (UKF) \cite{van2001unscented} and 
Particle Filtering (PF) \cite{2002Particle} in the backend of SLAM.
The above mentioned Hector-SLAM used the EKF algorithm framework, while FastSLAM and Gmapping adopted PF algorithm architecture.
Later studies began to explore optimization-based SLAM algorithm and achieved better performance.
The KartoSLAM, LagoSLAM, LOAM, LeGO-LOAM, IMLS-SLAM, Suma, and Suma++ all used the optimization algorithm 
in their backends.
Our previous study also utilized the optimization method to build a pose graph in a LiDAR-SLAM system \cite{chen2019trajectory}.

Research concerning Global Navigation Satellite System (GNSS) 
has lasted for decades since the advent of 
Global Positioning System (GPS). 
GNSS includes GPS, GLONASS, Galileo and BeiDou. 
Its operation modes mainly involve four types: 
Single Point Positioning (SPP), 
Real-Time Differential (RTD) Positioning, 
Real-Time Kinematic (RTK) Positioning \cite{langley1998rtk} and Precise Point Positioning (PPP) \cite{1997Precise}. 
The positioning accuracy would be affected by many factors including receiver clock errors, 
satellite clock errors, tropospheric delays, ionospheric delays, hardware code and phase biases, 
multipath errors, tide, antenna phase center, phase winding, 
earth rotation, relativity effects, etc. 
The ionospheric error is of significance for single-frequency GNSS users \cite{xiang2017carrier}.

The characteristics of the four GNSS operation modes will be introduced below.
SPP uses pseudorange and Doppler observation to locate and measure velocity. 
Due to the heavy noise of pseudorange itself, the SPP generally fixes errors above the meter level, 
including satellite clock errors, relativistic effects, tropospheric delay, ionospheric errors, and earth rotation.
Generally, in SPP, the tropospheric error is corrected by empirical models like Hopfield \cite{hopfield1969two} and Saastamoinen \cite{saastamoinen1973contributions}; 
the ionospheric error is modified by Klobuchar model \cite{klobuchar1987ionospheric}; 
the satellite orbit error and clock error are both corrected by broadcast ephemeris; 
relativistic effects are corrected according to the model; 
earth rotation errors can be corrected according to the signal propagation time.
RTD uses reference stations with known fixed positions to calculate 
the differences between the measured pseudoranges and actual pseudoranges. 
The differences are then broadcast to nearby mobile stations for use.
RTK corrects not only the pseudorange observation errors but also the phase observation errors, 
so as to obtain the positioning accuracy of centimeter level. 
RTD and RTK both need base stations. 
In contrast, PPP technology is more suitable for unmanned systems as PPP is independent of base stations.
Single-frequency PPP (SF-PPP) and Dual-frequency PPP (DF-PPP) are two mainstream PPP technologies.
Both of them need precise corrections which can be acquired from International GNSS Service Real-Time Service (IGS-RTS) \cite{IGS-RTS} for the satellite orbits and satellite clocks.
In autonomous driving, SF-PPP receives considerable attention because of its lower cost. 
However, the positioning performance of any GNSS modes will be affected by 
the coarse errors introduced by multipath and non-line-of-sight (NLOS) signals \cite{9184896}. 
The observations with coarse errors are defined as GNSS outliers which can be eliminated by 
GNSS Receiver Autonomous Integrity Monitoring (RAIM) \cite{kalafus1987receiver}.
In recent years, multi-sensors assisted RAIM algorithms have been proposed \cite{gong2019tightly}, \cite{zhu2019dual}.
For example, inertial measurement unit (IMU) was used in RAIM with a tightly-coupled GNSS/INS(Inertial Navigation System) framework \cite{hewitson2010extended}.
Our previous work also used LiDAR to analyze line-of-sight GNSS observations \cite{Hyyppa2017Feasibility}.

In autonomous driving, LiDAR-SLAM registers the point cloud to obtain the transformation 
matrix between two frames, then the trajectory is obtained, and maps are built accordingly. 
In this process, the error of LiDAR-SLAM inevitably occurs along with the distance drift. 
And in the open areas with limited features, LiDAR-SLAM may fail to work while GNSS performs well.
Moreover, it is impossible to estimate the position in the earth coordinate system 
if we simply use the LiDAR-SLAM technology. This makes some navigation tasks more challenging.
For these reasons, the coupling of GNSS will enable the LiDAR-SLAM system to provide more robust 
localization and mapping results with global earth coordinates in large-scale scenarios.
In turn, GNSS is heavily influenced by multipath and NLOS in urban canyon environment, 
resulting in poor performance. 
LiDAR-SLAM always performs well in urban canyon environment and gets a relative positioning result.
So it is meaningful to fuse GNSS and LiDAR-SLAM: LiDAR-SLAM can improve the reliability and availability of GNSS, 
on the other hand, GNSS can turn the positioning result of LiDAR-SLAM in a local coordinate system to the global such as 
World Geodetic System 1984 (WGS-84).
But in conventional GNSS and LiDAR-SLAM fusion systems, few works have paid attention to 
either the calculation of LiDAR-SLAM positioning covariance, or 
the use of PPP in autonomous driving. 
In conclusion, our motivation for this paper is complementing PPP and LiDAR-SLAM to obtain more reliable 
and available positioning results in global coordinate system.

The main contributions of this paper are as follows:

1) Covariance is essential for weighting different positioning sources in a multi-sensor integrated localization approach.
However, few LiDAR-SLAM systems provide covariance estimation since only one sensor is utilized. 
For more accurate integration of LiDAR-SLAM and PPP, 
we derive an SVD Jacobian based error propagation model to estimate the covariance of LiDAR-SLAM. 
The estimated result is evaluated by a novel method we propose, using GNSS covariance and ground truth of trajectory.
The evaluation proves that the estimated LiDAR-SLAM covariance is reasonable.

2) Urban canyon environment always introduces GNSS observations with more coarse errors 
which may result in the failure of RAIM.
In such an environment, LiDAR might provide more stable positioning results.
Therefore, we propose a LiDAR-SLAM assisted GNSS RAIM to eliminate PPP outliers, achieving
reliable GNSS positioning results. The max error of PPP is reduced to the same level as LiDAR-SLAM.

3) By applying the above SVD Jacobian based error propagation model and LiDAR-SLAM assisted GNSS RAIM, 
we develop a novel loosely coupled SLAM system called P$^3$-LOAM.
Then P$^3$-LOAM was evaluated by UrbanNav, a challenging public dataset in urban canyon environment.
P$^3$-LOAM significantly outperforms benchmarks in terms of accuracy and availability.

The remaining paper is organized as follows. 
Section \Rmnum{2} presents an overview of related work. 
Summary of the proposed architecture and algorithm is provided in Section \Rmnum{3}. 
Section \Rmnum{4} describes the experimental setup and presents the results. 
Finally, conclusions are drawn in Section \Rmnum{5}.

\section{Related work}
Lots of studies have integrated GNSS and LiDAR-SLAM 
for better localization performance.
LiDAR-SLAM positioning covariance was considered 
in Shetty et al.'s work \cite{Akshay2019Adaptive}, 
by modeling the positioning error as a Gaussian distribution 
and estimating the covariance 
from individual surface and edge feature points. 
This paper hypothesized that each surface feature point 
contributed to reducing positioning error in the direction of 
the corresponding surface normal, while 
each edge feature point helped in reducing position error 
in the direction perpendicular to the edge vector.
But the eigenvalues of covariance matrix were fixed in this algorithm, 
so it is not adaptive for real-world circumstances where 
the eigenvalues would change with distance.
Shetty et al. used Kalman Filtering to fuse LiDAR-SLAM and 
GNSS \cite{Akshay2019Adaptive}. For transforming LiDAR-SLAM position to
the global frame, their paper used an on-board 
attitude and heading reference system (AHRS) combined with an IMU.
Constant velocity model was chosen as the dynamic model of Kalman Filtering, 
but the error propagation model of IMU would 
be a better choice in this case. 
Similarly, Pereira et al. \cite{Antonio2019GNSS} used GNSS, 
LiDAR and AHRS to navigate through the forest by constructing UKF.
In contrast, IMU was used sufficiently through IMU 
preintegration \cite{Chang2019GNSS}. 
However, its backend was not filtering but optimization method, 
and the covariance of LiDAR-SLAM was not considered. 
In addition, the LiDAR-SLAM algorithm in \cite{Chang2019GNSS} was Cartographer, 
which is designed for indoor environment instead of urban canyon environment.
We choose LOAM series LiDAR-SLAM algorithm in our paper as 
it obtains better localization performance on KITTI \cite{geiger2012we}. 
Qian et al. \cite{qian2020lidar} used EKF to fuse RTK, IMU, and 2D LiDAR. 
They proved LiDAR could improve RTK ambiguity resolution. 
Qian et al argued that IMU could only keep high accuracy for a short time 
because of its drift errors while LiDAR could perform better.
Similarly, 3D LiDAR was also fused with RTK and IMU using EKF in \cite{wan2018robust}. 
In the works above, IMU is crucial in converting global and local coordinates, 
but it is actually not a necessary component, 
as exemplified in \cite{joerger2006autonomous}, \cite{he2020integrated}, 
\cite{kanhere2018integrity} and \cite{Weisong2020GNSS}.
Joerger et al. \cite{joerger2006autonomous} used Kalman Filtering to fuse 2D LiDAR and GPS, 
resulting in a robust localization in forest scenario and urban canyon scenario.
It is worth mentioning that Carrier-Phase Differential GPS was used in \cite{joerger2006autonomous}. 
Shamsudin et al. \cite{2018Consistent} extended Rao-Blackwellized particle filtering to fuse RTK and 3D LiDAR. 
He et al. \cite{he2020integrated} used an optimization-based
algorithm to integrate RTK and 3D LiDAR in partially GNSS-denied environment. 
In \cite{kanhere2018integrity}, the authors used a RAIM framework to integrate LiDAR odometry and GNSS by UKF. 
Hsu et al. \cite{Weisong2020GNSS} focused on GNSS covariance and applied Gaussian Mixture Model (GMM) 
to identify non-Gaussian GNSS outliers in SPP algorithm.
The results were satisfactory by integrating SPP and LiDAR-SLAM, 
but there was still room for improvement.
We therefore propose to utilize PPP technology and 
consider LiDAR-SLAM covariance in search for better results.
\begin{figure*}[htp]
  \centering
    \begin{center}
      \includegraphics[width=14.28cm,height=7.4311cm]{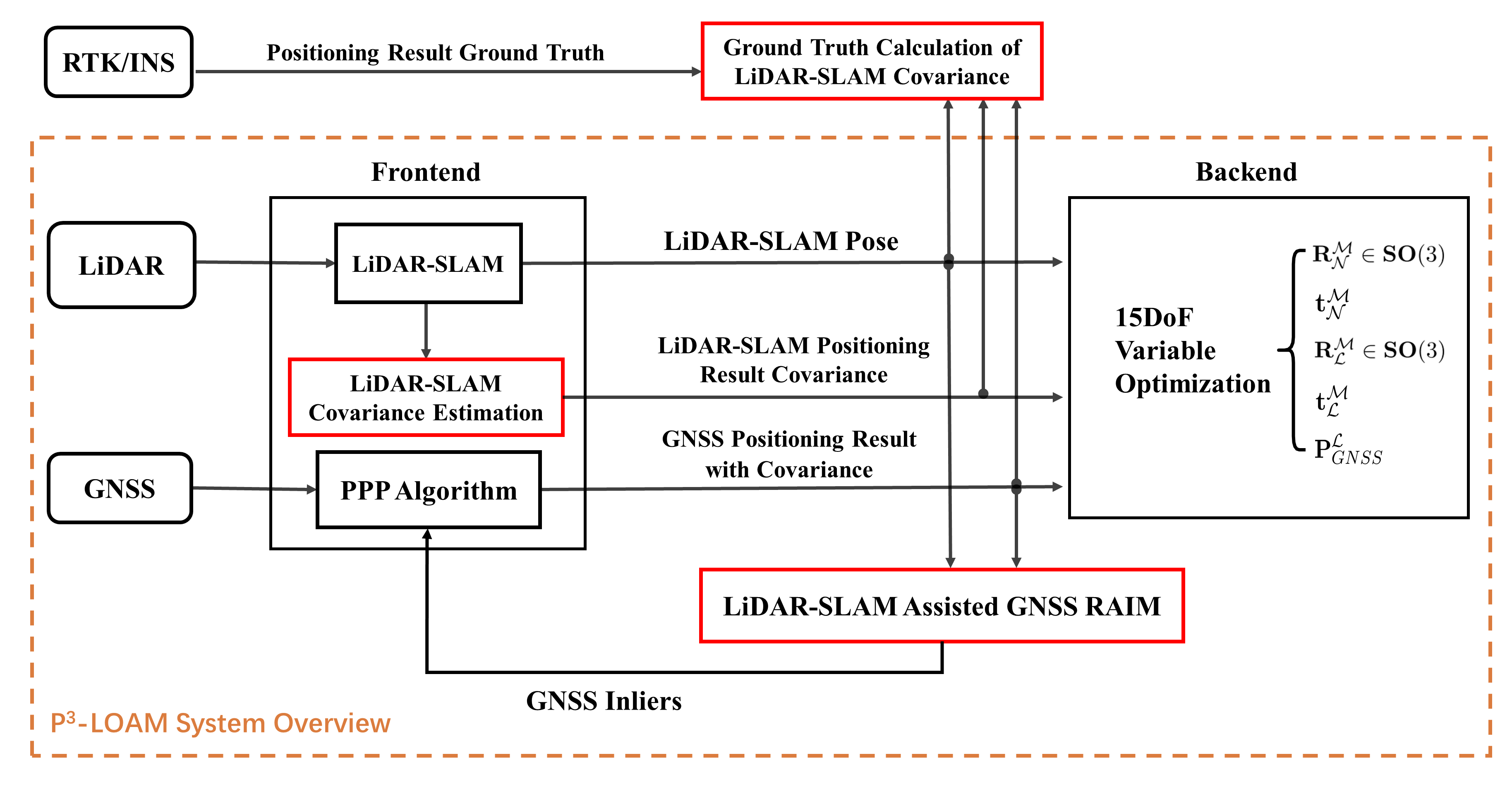}
    \end{center}
       \caption{Overview of the Modules Involved}  
    \label{fig:SystemOverview}
\end{figure*}

\section{Methodology}

In this section, the specific implementation of P$^3$-LOAM will be presented first.
Secondly, we will briefly introduce the ICP algorithm and derive 
its error propagation based on SVD Jacobian model.
Thirdly, the implementation of PPP algorithm and its covariance will be explicated.
On this basis, the calculation method of LiDAR-SLAM covariance's ground truth will be derived. 
Next, the LiDAR-SLAM assisted GNSS RAIM algorithm will also be proposed.
At last, we will build up the factor graph model of the backend.

\subsection{Overview}
All the modules involved in this paper are presented in Fig. \ref{fig:SystemOverview}. 
The content within the dashed box is the general architecture of the proposed P$^3$-LOAM, 
and the outside part is a method to calculate the ground truth of LiDAR-SLAM covariance. 
P$^3$-LOAM is divided into frontend and backend.
In the frontend, LiDAR-SLAM and PPP algorithm are the two main modules. 
In the LiDAR-SLAM module, we add a covariance estimation. 
In the PPP module, we propose a RAIM to select GNSS inliers with the help of LiDAR-SLAM pose and PPP result. 
At last, a backend with 15DoF variable optimization is realized. 
The parts in Fig. \ref{fig:SystemOverview} with red rectangles represent the contributions of our work.

\begin{figure}[htp]
  \centering
    \begin{center}
    \includegraphics[width=8.8cm,height=5.75cm]{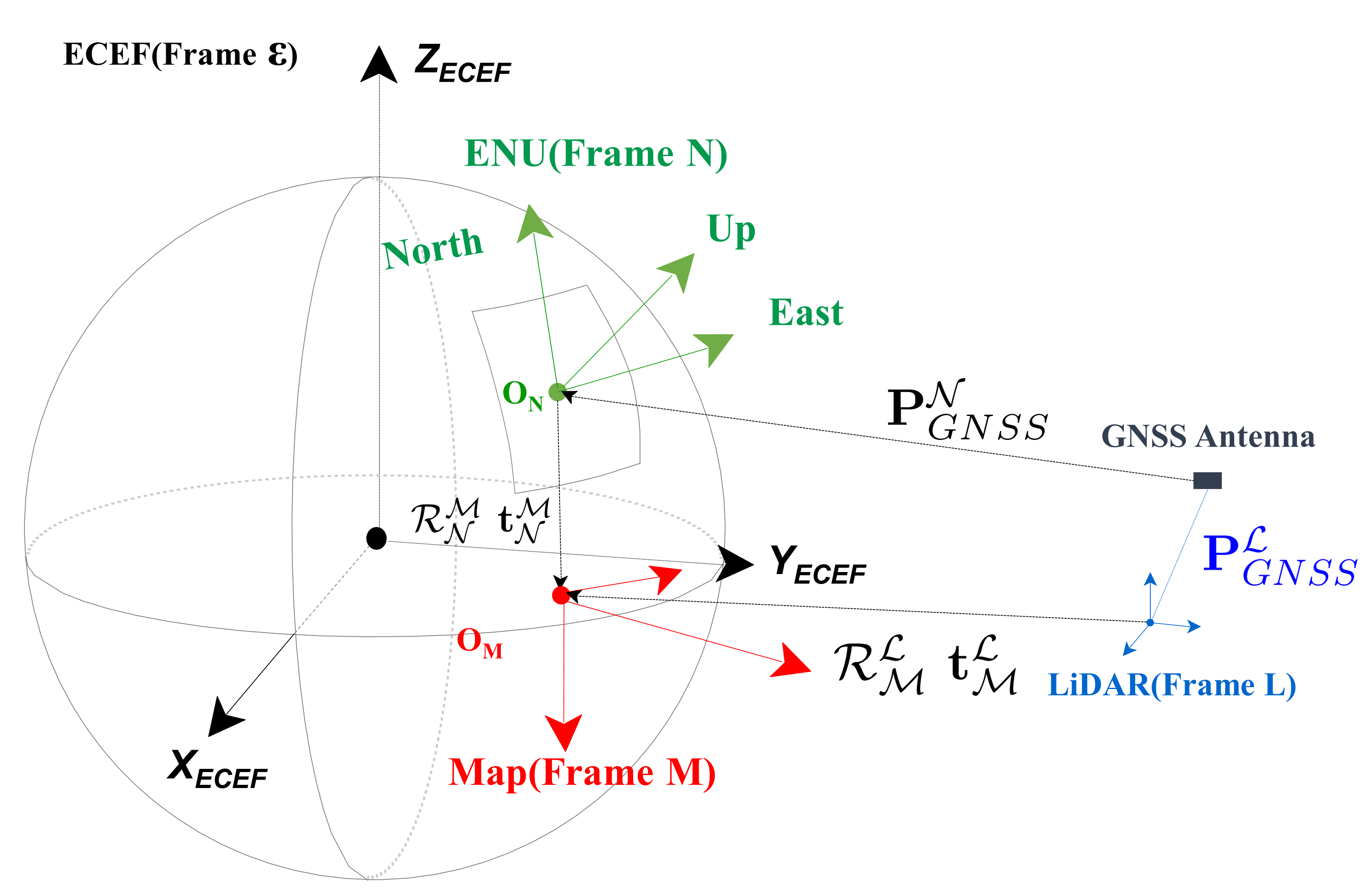}
  \end{center}
       \caption{Coordinate Frames}
    \label{fig:CoordinateFrame}
\end{figure}

The coordinate frames and notations involved in this article will be clarified below.
The Earth-centered Earth-fixed coordinate system (ECEF, Frame $\mathcal{E}$, see Fig. \ref{fig:CoordinateFrame}) 
rotates with the Earth, taking the Earth's centroid as the origin. The X-axis of Frame $\mathcal{E}$ points to 
the intersection of the equator and prime meridian. The Earth's rotation axis is taken as Z-axis, 
and the North Pole is the positive direction. Then, the Y-axis is perpendicular to the X-Z plane, forming a right-handed coordinate system.
The GNSS receiver generally outputs positioning results in Frame $\mathcal{E}$. 

The East-North-Up coordinate system (ENU) is marked as Frame $\mathcal{N}$. 
When Frame $\mathcal{N}$ is used, we have to choose a point on the Earth as the origin. 
And the X-Y plane of Frame $\mathcal{N}$ is the local horizontal plane of the origin, 
with the X-axis being the tangential line of the latitude line that points East, 
and the Y-axis being the tangential line of the longitude line that points North. 
The Z-axis is perpendicular to the X-Y plane, forming a right-handed coordinate system.

The LiDAR frame is marked as Frame $\mathcal{L}$, with LiDAR's center being the origin.
The X-axis points to the right along the LiDAR horizontal axis, the Z-axis points forward along the LiDAR longitudinal axis, 
and the Y-axis is perpendicular to the X-Z plane, forming a right-handed coordinate system.

A local map frame is marked as Frame $\mathcal{M}$, coinciding with Frame $\mathcal{L}$ at the beginning and then keeps still.

For convenience of reading, we summarize all the involved coordinate frames and notations in Table \ref{table:Notations}.
\begin{table}[htp]
  \centering
  \caption{Coordinate Frames and Notations}
  \setlength{\tabcolsep}{0.8mm}{
  \begin{tabular*}{250pt}{cccc}
  \hline
  \hline
  $\mathcal{E}$& ECEF &$\mathcal{N}$ &ENU Frame \\ 

  \hline
  $\mathcal{L}$& LiDAR Frame &$\mathcal{M}$ &Locam Map Frame \\ 

  \hline
  $\mathbf{R}_{\mathcal{A}}^{\mathcal{B}}$&\makecell[c]{Rotation Matrix from \\Frame $\mathcal{A}$ to Frame $\mathcal{B}$}
  &\makecell[c]{$\delta \theta_{\mathcal{A}}^{\mathcal{B}}$}&\makecell[c]{Small disturbance 
  \\to the left of $\mathbf{R}_{\mathcal{A}}^{\mathcal{B}}$}\\ [7pt]
  \hline
  $\mathbf{t}_{\mathcal{A}}^{\mathcal{B}}$&\makecell[c]{Translation Vector from \\Frame $\mathcal{A}$ to Frame $\mathcal{B}$}  
  & $\delta \mathbf{t}_{\mathcal{A}}^{\mathcal{B}}$  & \makecell[c]{Small disturbance 
  \\to the left of $\mathbf{t}_{\mathcal{A}}^{\mathcal{B}}$} \\[7pt]
  \hline
  $\mathbf{P}_{\mathbf{G}}^{\mathcal{A}}$&\makecell[c]{The point $\mathbf{G}$'s \\coordinates in Frame $\mathcal{A}$}  & 
  \makecell[c]{$Cov(\mathbf{X})$\\or $\mathbf{\Lambda_{X}}$}&\makecell[c]{Covariance of $\mathbf{X}$}  \\
  \hline
  $\mathbf{U}_{jk}$&\makecell[c]{The $j$th row and the $k$th \\column of matrix $\mathbf{U}$}  
  & $\mathbf{\Omega}$$^{ij}_{\mathbf{U}_{kl}}$  & \makecell[c]{The $k$th row and the $l$th \\column of matrix
  $\mathbf{\Omega}$$^{ij}_{\mathbf{U}}$} \\
  \hline
  $\mathbf{P,Q}$&\makecell[c]{Two point sets \\in polar coordinates\\(LiDAR original observation)}
  & $\overrightarrow{X_{A} X_{B}}$  & \makecell[c]{Vector from point A to B} \\
  \hline
  \hline
  \end{tabular*}}
  \label{table:Notations}
\end{table}

\subsection{Covariance Estimation of LiDAR-SLAM}
In many mainstream LiDAR-SLAM systems, ICP is a key issue which can be solved by many methods to obtain the rotation and translation
between two point clouds. 
After this, the corresponding relations between these two point clouds are obtained, then we use SVD Jacobian model to analyze the covariance of rotation and translation.
\subsubsection{An explicit analytic solution of ICP using SVD}

There are two point sets,
\{$\mathbf{A}_i\}_{i=1}^{n}$ and \{$\mathbf{B}_i\}_{i=1}^{n}$.
Their centers are marked as $\{\mathbf{A}_c\}$ and $\{\mathbf{B}_c\}$.
Both of them can be calculated by Eq.(\ref{equ:AcBc}).
\begin{equation}
\label{equ:AcBc}
    \begin{cases}
        \mathbf{A}_c=\frac{1}{n} \sum_{i=1}^{n}\mathbf{A}_i \\
        \mathbf{B}_c=\frac{1}{n} \sum_{i=1}^{n}\mathbf{B}_i
    \end{cases}
\end{equation}

\{$\mathbf{M}_i\}_{i=1}^{n}$ and \{$\mathbf{N}_i\}_{i=1}^{n}$ are the normalized coordinates of \{$\mathbf{A}_i\}_{i=1}^{n}$
and \{$\mathbf{B}_i\}_{i=1}^{n}$. They can be calculated by Eq.(\ref{equ:MiNi}).
\begin{equation}
\label{equ:MiNi}
    \begin{cases}
        \mathbf{M}_i=\mathbf{A}_i-\mathbf{A}_c \\
        \mathbf{N}_i=\mathbf{B}_i-\mathbf{B}_c
    \end{cases}
\end{equation}

Matrix $\mathbf{W}$ can be calculated by Eq.(\ref{equ:W}).
\begin{equation}
\label{equ:W}
    \mathbf{W}=\sum_{i=1}^{n}\mathbf{N}_i\mathbf{M}_i^T
\end{equation}

We apply SVD to $\mathbf{W}$ using Eq.(\ref{equ:Wsvd1}).
$\mathbf{U}$ and $\mathbf{V}$ are both 3$\times$3 orthonormal matrices.
The columns of $\mathbf{U}$ and $\mathbf{V}$ are defined as  
the left-singular vectors and right-singular vectors of $\mathbf{W}$, respectively.
$\mathbf{D}$ is a 3$\times$3 diagonal matrix with the singular values of $\mathbf{W}$.
\begin{equation}
\label{equ:Wsvd1}
    \mathbf{W}=\mathbf{UDV}^T
\end{equation}

Then the rotation matrix and the translation vector from \{$\mathbf{A}_i\}_{i=1}^{n}$ 
to \{$\mathbf{B}_i\}_{i=1}^{n}$ can be calculated by Eq.(\ref{equ:Rt}).
\begin{equation}
    \begin{cases}
    \label{equ:Rt}
    \mathbf{R}_{\mathcal{A}}^{\mathcal{B}}=\mathbf{UV}^{T} \\
    \mathbf{t}_{\mathcal{A}}^{\mathcal{B}}=\mathbf{B}_c-\mathbf{R}_{\mathcal{A}}^{\mathcal{B}}\mathbf{A}_c
    \end{cases}
\end{equation}

\subsubsection{ICP Error Propagation Model based on SVD Jacobian}
The error propagation model of ICP is proposed in this subsection based on SVD Jacobian.
According to \cite{papadopoulo2000estimating}, the
Jacobian matrices of $\mathbf{U}$, $\mathbf{D}$ and $\mathbf{V}$ with respect to $\mathbf{W}$ 
can be calculated by Eq.(\ref{equ:Wjacobian1}).
\begin{equation}
\label{equ:Wjacobian1}
    \begin{cases}
    \frac{\partial \mathbf{D}_{kk}}{\partial \mathbf{W}_{ij}} = \mathbf{U}_{ik}\mathbf{V}_{jk} \\
    \frac{\partial \mathbf{U}}{\partial \mathbf{W}_{ij}} = \mathbf{U\Omega}_{\mathbf{U}}^{ij} \\
    \frac{\partial \mathbf{V}}{\partial \mathbf{W}_{ij}} = -\mathbf{V\Omega}_{\mathbf{V}}^{ij}
    \end{cases}
\end{equation}

$\mathbf{\Omega}$$^{ij}_{\mathbf{U}}$ and $\mathbf{\Omega}$$^{ij}_{\mathbf{V}}$ are both matrices and their elements 
$\mathbf{\Omega}$$^{ij}_{\mathbf{U}_{kl}}$ and $\mathbf{\Omega}$$^{ij}_{\mathbf{V}_{kl}}$can be caluculated by Eq.(\ref{equ:OmegaUV1}).
\begin{equation}
    \label{equ:OmegaUV1}
        \begin{cases}
            \mathbf{D}_{ll}\mathbf{\Omega}_{\mathbf{U}_{kl}}^{ij}+\mathbf{D}_{kk}\mathbf{\Omega}_{\mathbf{V}_{kl}}^{ij} = \mathbf{U}_{ik}\mathbf{V}_{jl} \\
            \mathbf{D}_{kk}\mathbf{\Omega}_{\mathbf{U}_{kl}}^{ij}+\mathbf{D}_{ll}\mathbf{\Omega}_{\mathbf{V}_{kl}}^{ij} = -\mathbf{U}_{il}\mathbf{V}_{jk}
        \end{cases}
\end{equation}

According to Eq.(\ref{equ:Rt}), the covariance of $\mathbf{R}_{\mathcal{A}}^{\mathcal{B}}$,
marked as $\mathbf{\Lambda_{R^\mathcal{B}_\mathcal{A}}}$, can be calculated by Eq.(\ref{equ:lambdaR}).

\begin{equation}
  \label{equ:lambdaR}
      \begin{cases}
      \mathbf{\Lambda_{R^\mathcal{B}_\mathcal{A}}}=\frac{\partial \mathbf{UV}^{T}}{\partial \mathbf{W}}\mathbf{\Lambda_{W}}\frac{\partial (\mathbf{UV}^{T})^{T}}{\partial \mathbf{W}}\\
      \frac{\partial \mathbf{UV}^{T}}{\partial \mathbf{W}_{ij}}=\frac{\partial \mathbf{U}}{\partial \mathbf{W}_{ij}}\mathbf{V}^{T}+\mathbf{U}\frac{\partial \mathbf{V}^{T}}{\partial \mathbf{W}_{ij}}
      \end{cases}
  \end{equation}

After $\mathbf{\Lambda_{R^\mathcal{B}_\mathcal{A}}}$ is obtained, $\mathbf{\Lambda_{t^\mathcal{B}_\mathcal{A}}}$ 
can be calculated by Eq.(\ref{equ:lambdat}).
\begin{equation}
\label{equ:lambdat}
    \begin{cases}
        \mathbf{\Lambda_{t^\mathcal{B}_\mathcal{A}}}=\mathbf{\Lambda}_{\mathbf{B}_c}+\mathbf{\Lambda}_{\mathbf{R}_{\mathcal{A}}^{\mathcal{B}}\mathbf{A}_c} \\
        \mathbf{\Lambda}_{\mathbf{R}_{\mathcal{A}}^{\mathcal{B}}\mathbf{A}_c}=\mathbf{A}_c^T \mathbf{\Lambda}_{\mathbf{R}_{\mathcal{A}}^{\mathcal{B}}} \mathbf{A}_c
    \end{cases}
\end{equation}

The original measurements of \{$\mathbf{A}_i\}_{i=1}^{n}$ and \{$\mathbf{B}_i\}_{i=1}^{n}$ 
are in polar coordinates in the form of one distance and two angles, marked as $\mathbf{d}$ (distance between LiDAR and the measured point), 
$\bm{\omega}$ (vertical angle) and $\bm{\alpha}$ (horizontal angle).
The original measurement error in polar coordinates can be passed to Cartesian coordinates by coordinate transformation so that we can 
easily obtain $\mathbf{\Lambda_{W}}$ in Eq.(\ref{equ:lambdaR}).

In conclusion, we summarize the ICP error propagation model based on SVD Jacobian in Algorithm \ref{alg:ICP Error Propagation Model based on Jacobian of SVD}.
\begin{algorithm}[htb]  
  \caption{ICP Error Propagation Model based on SVD Jacobian}  
  \label{alg:ICP Error Propagation Model based on Jacobian of SVD}  
  \begin{algorithmic}[1]  
    \Require  
    Two point sets in polar coordinates $\mathbf{P,Q}$ and their covariance $\mathbf{\Lambda_{P}}$, $\mathbf{\Lambda_{Q}}$; 
    ICP results between $\mathbf{P}$ and $\mathbf{Q}$:
    $\mathbf{t}_{\mathbf{P}}^{\mathbf{Q}}$, $\mathbf{R}_{\mathbf{P}}^{\mathbf{Q}}$;
    Corresponding relations between $\mathbf{P}$ and $\mathbf{Q}$: $\mathbf{C}_{\mathbf{PQ}}$; 
    \Ensure  
    $\mathbf{\Lambda_{t^\mathbf{Q}_\mathbf{P}}}$, $\mathbf{\Lambda_{R^\mathbf{Q}_\mathbf{P}}}$;
    \State [$\mathbf{A,B}$, $\mathbf{\Lambda_{A}}$, $\mathbf{\Lambda_{B}}$]=Polar2Cart($\mathbf{P,Q}$, $\mathbf{\Lambda_{P}}$, $\mathbf{\Lambda_{Q}}$);
    //Turn the point sets coordinates and their covariance from polar coordinates to Cartesian coordinates;

    \State $\mathbf{t}_{\mathbf{A}}^{\mathbf{B}}$ = $\mathbf{t}_{\mathbf{P}}^{\mathbf{Q}}$, 
    $\mathbf{R}_{\mathbf{A}}^{\mathbf{B}}$ = $\mathbf{R}_{\mathbf{P}}^{\mathbf{Q}}$,
    $\mathbf{C}_{\mathbf{AB}}$ = $\mathbf{C}_{\mathbf{PQ}}$;

    \State Eq.(\ref{equ:AcBc}), (\ref{equ:MiNi}), (\ref{equ:W}), (\ref{equ:Wsvd1}) to calculate $\mathbf{U,D,V}$ and $\mathbf{\Lambda_{W}}$;

    \State Eq.(\ref{equ:Wjacobian1}), (\ref{equ:OmegaUV1}) to calculate the Jacobian matrix of $\mathbf{U,D,V}$ with respect to $\mathbf{W}$;

    \State Eq.(\ref{equ:lambdaR}) to calculate $\mathbf{\Lambda_{t^\mathbf{B}_\mathbf{A}}}$;
    
    \State Eq.(\ref{equ:lambdat}) to calculate $\mathbf{\Lambda_{\mathbf{R}^\mathbf{B}_\mathbf{A}}}$;

    \State $\mathbf{\Lambda_{t^\mathbf{Q}_\mathbf{P}}}$ = $\mathbf{\Lambda_{t^\mathbf{B}_\mathbf{A}}}$
    $\mathbf{\Lambda_{\mathbf{R}^\mathbf{Q}_\mathbf{P}}}$ = $\mathbf{\Lambda_{\mathbf{R}^\mathbf{B}_\mathbf{A}}}$;  
    \\  
    \Return $\mathbf{\Lambda_{t^\mathbf{Q}_\mathbf{P}}}$, $\mathbf{\Lambda_{R^\mathbf{Q}_\mathbf{P}}}$;  
  \end{algorithmic}  
\end{algorithm}

\subsection{Single Frequency Precise Point Positioning}
In our paper, we choose SF-PPP algorithm to realize PPP for two reasons.
One is that single-frequency GNSS receivers are still favored by a large number of users because of its low cost. 
The other is that the dataset we use only includes single-frequency GNSS data.
The basic SF-PPP observation equations are listed in Eq.(\ref{equ:SFPPP}) \cite{chen2005real}.

\begin{equation}
\begin{split}
    \label{equ:SFPPP}
    \begin{cases}
      \operatorname{Pr} = p_{r}^{s} + c\cdot\left(dt_{r}-dT^{s}\right) + d_{orb} + d_{ion} + d_{trop}\\
      \qquad + d_{rel} + \varepsilon(\operatorname{Pr}) \\
      \Phi_{L} = p_{r}^{s} + c\cdot\left(dt_{r} - dT^{s}\right) + d_{orb} - d_{ion} + d_{trop} \\
      \qquad + \lambda_{L} \cdot N_{L}+d_{rel} + d_{pw} + \varepsilon\left(\Phi_{L}\right)  
    \end{cases}
\end{split}
\end{equation}

$\operatorname{Pr}$ is pseudorange measurement while 
$\Phi_{L}$ stands for carrier phase range measurement.
$p_{r}^{s}$ is the true geometric range between the GNSS receiver and satellite.
$c$ means the speed of light. 
$d t_{r}$ represents the GNSS receiver clock error to be estimated.
$d T^{s}$ is the satellite clock error which can be eliminated by precise clock products, 
and $d_{orb}$ stands for the satellite orbit error eliminated by precise orbit products.
$d_{ion}$ is the ionospheric error, mitigated by using Global Ionosphere Maps(GIMs) in our paper.
$d_{trop}$ is the tropospheric error, corrected by Saastamoinen model.
$d_{rel}$ denotes the relativistic effects and $d_{pw}$ is the phase windup error 
on the carrier phase measurements.
$\lambda_{L}$ is the wavelength of the carrier phase.
$N_{L}$ is phase ambiguity.
$\varepsilon(\operatorname{Pr})$ is pseudorange measurement noise 
and $\varepsilon\left(\Phi_{L}\right)$ means carrier phase range measurement noise.
It is interesting to note here that Difference Code Bias (DCB) of 
P2/P1 and P1/C1 need to be corrected by IGS products for Ionosphere \cite{CODE}.

For every epoch and every GNSS satellite, we obtain Eq.(\ref{equ:SFPPP}), 
then the equations of all satellites in one epoch are linearized and marked as Eq.(\ref{equ:SFPPPs}).
$\mathbf{Z}_{C}$ and $\mathbf{Z}_{P}$ respectively represent 
the carrier phase range and pseudorange measurements of all GNSS satellites. 
$\mathbf{B}_{C}$ and $\mathbf{B}_{P}$ are the corresponding Jacobian matrices.
\begin{equation}
    \label{equ:SFPPPs}
    \begin{split}
    \left[
    \begin{smallmatrix}
      \mathbf{Z}_{C} \\
      \mathbf{Z}_{P} 
    \end{smallmatrix}
    \right] &=     
    \left[
    \begin{smallmatrix}
        \mathbf{B}_{C}\\
        \mathbf{B}_{P}
      \end{smallmatrix}
    \right]   
    \left[
    \begin{smallmatrix}
        \mathbf{X}\\
        c\cdot dtr\\
        \mathbf{N}_{L}
      \end{smallmatrix}
    \right]
     +    
    \left[
    \begin{smallmatrix}
        \mathbf{\Delta}_{C}\\
        \mathbf{\Delta}_{P}
    \end{smallmatrix}
    \right]
    \end{split}
\end{equation}

Since the pseudorange's variance $\mathbf{D(\Delta}_{P})$ is different from the carrier phase's variance $\mathbf{D(\Delta}_{C})$, 
the weights of pseudorange and carrier phase are different.
If there are \textit{n} satellites observed in one epoch, 
then $\mathbf{V}$ is the residual vector of this epoch, which can be calculated by Eq.(\ref{equ:V}).
\begin{equation}
    \label{equ:V}
    \begin{split}
      \mathbf{V} &=    
      \left[
    \begin{smallmatrix}
      \mathbf{Z}_{C} \\
      \mathbf{Z}_{P} 
    \end{smallmatrix}
    \right] - 
    \left[
    \begin{smallmatrix}
        \mathbf{B}_{C}\\
        \mathbf{B}_{P}
      \end{smallmatrix}
    \right]   
    \left[
    \begin{smallmatrix}
        \mathbf{\tilde{X}}\\
        c\cdot \tilde{dtr}\\
        \mathbf{\tilde{N}}_{L}
      \end{smallmatrix}
    \right]=\mathbf{Z} - \mathbf{B}\left[
        \begin{smallmatrix}
            \mathbf{\tilde{X}}\\
            c\cdot \tilde{dtr}\\
            \mathbf{\tilde{N}}_{L}
          \end{smallmatrix}
        \right]
    \end{split}
\end{equation}

SF-PPP covariance, $cov(\mathbf{\tilde{X}})$, is the first three rows and first three columns 
of Eq.(\ref{equ:covGNSS}),
$cov(\mathbf{\tilde{X}},c\cdot \tilde{dtr},\mathbf{\tilde{N}}_{L})_{(1:3,1:3)}$.
\begin{small} 
\begin{equation}
    \label{equ:covGNSS}
    \begin{split}
    cov(\mathbf{\tilde{X}},c\cdot \tilde{dtr},\mathbf{\tilde{N}}_{L})
    &=
    \frac{\mathbf{V}^{\mathrm{T}}
    \mathbf{P}
    \mathbf{V}}{n-4}\left(\mathbf{B}^{\mathrm{T}} 
    \mathbf{P}
    \mathbf{B}\right)^{-1}
    =m_{0}^{2} \cdot \mathbf{Q}_{\mathrm{x}}\\
    P&=\left[
      \begin{smallmatrix}
          \mathbf{D(\Delta}_{C})&0\\
          0&\mathbf{D(\Delta}_{C})\\
        \end{smallmatrix}
      \right]^{-1}
    \end{split}
\end{equation}
\end{small}

The diagonal of $\mathbf{Q}_{\mathrm{x}}$ constitutes the DOP (Dilution of Precision) values 
which reflect the spatial structure of the satellites.

\subsection{Ground Truth Calculation of LiDAR-SLAM Positioning Covariance Assisted with GNSS}
\begin{figure}[htp]
    \begin{center}
    \includegraphics[width=8cm,height=3.319cm]{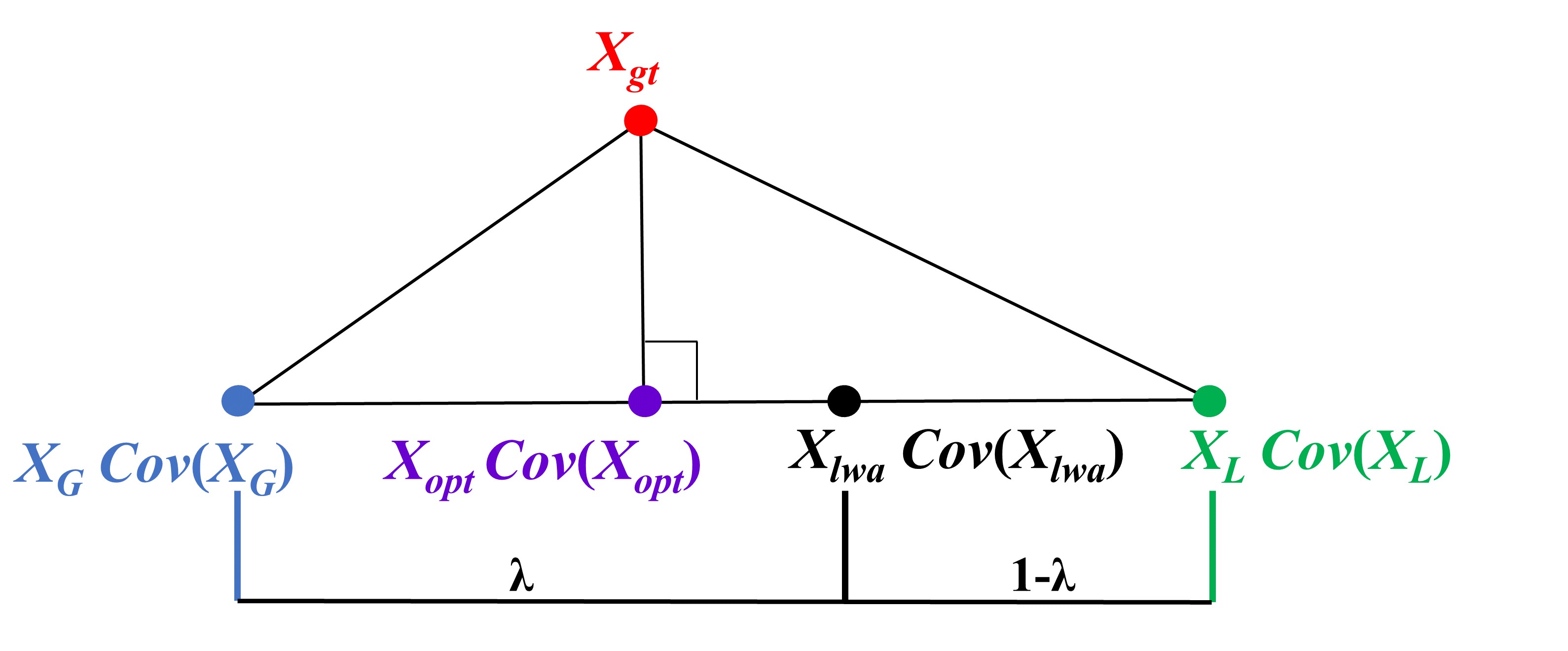}
    \end{center}
       \caption{Ground Truth Calculation of LiDAR-SLAM Position Covariance Assisted with GNSS}
    \label{fig:cal LiDAR cov by GNSS cov}
\end{figure}

Based on GNSS positioning covariance introduced above, 
we calculate the ground truth of LiDAR-SLAM positioning covariance in this subsection.
There are three positioning results in Fig. \ref{fig:cal LiDAR cov by GNSS cov}: the red $X_{g t}$ is ground truth,
the blue $X_{G}$ is GNSS positioning result and the green $X_{L}$ is LiDAR-SLAM positioning result.
$X_{G}$ and $X_{L}$ are independent of each other and both satisfy the Gaussian distribution. 
As such, if using linear weighted average algorithm to fuse $X_{G}$ and $X_{L}$, 
then the optimal estimation under linear weighted criterion, $X_{opt}$, must satisfy $X_{opt}X_{gt}\bot X_{G}X_{L}$. 
Through this vertical relationship, $X_{opt}$ can be calculated by Eq.(\ref{equ:opt}).
\begin{equation}
  \begin{split}
    \label{equ:opt}
    \overrightarrow{X_{G} X_{o p t}}
    &=
    \frac{\overrightarrow{X_{G} X_{g t}} \cdot 
    \overrightarrow{X_{G} X_{L}}}{|\overrightarrow{X_{G} X_{L}}|} 
    \cdot \frac{\overrightarrow{X_{G} X_{L}}}{|\overrightarrow{X_{G} X_{L}}|}\\
    \Rightarrow
    X_{o p t}
    &=
    \frac{\overrightarrow{X_{G} X_{g t}} \cdot 
    \overrightarrow{X_{G} X_{L}}}{|\overrightarrow{X_{G} X_{L}}|} 
    \cdot \frac{\overrightarrow{X_{G} X_{L}}}{|\overrightarrow{X_{G} X_{L}}|} + X_{G} 
  \end{split}
\end{equation}

After obtaining $X_{o p t}$, the ground truth of LiDAR-SLAM positioning covariance  
can be calculated by Eq.(\ref{equ:covXLiDAR}).

\begin{equation}
    \label{equ:covXLiDAR}
    Cov\left(X_{L}\right)=
    \frac{|\overrightarrow{X_{o p t} X_{L}}|}{|\overrightarrow{X_{G} X_{o p t}}|} 
    Cov\left(X_{G}\right) 
\end{equation}

Eq.(\ref{equ:covXLiDAR}) is proved as follows.
Since $X_L$ and $X_G$ are random variables subject to Gaussian distribution,
we mark them as $X_L\sim N(u_{X_L},Cov(X_L))$ and $X_G\sim N(u_{X_G},Cov(X_{G}))$.
$X_{lwa}$, the linear weighted average of $X_L$ and $X_G$, equals to 
$\lambda X_L + (1-\lambda)X_G$ and is subject to Eq.(\ref{equ:XYCOV}).

\begin{scriptsize} 
\begin{equation}
  \label{equ:XYCOV}
  X_{lwa} \sim N(\lambda u_{X_L}+(1-\lambda)u_{X_G},\lambda ^2Cov(X_L)+(1-\lambda)^2Cov(X_G))
\end{equation}
\end{scriptsize}

In addition, under linear weighted criterion, 
the weight is inversely proportional to the distance, 
so $\lambda$ must satisfy Eq.(\ref{equ:lambdageometric}).

\begin{equation}
  \label{equ:lambdageometric}
  \lambda |\overrightarrow{X_{G} X_{lwa}}| = (1-\lambda)|\overrightarrow{X_{lwa} X_{L}}|
\end{equation}

It is noted that the optimal value of $X_{lwa}$ is $X_{opt}$, 
which can be derived from minimizing $Cov(X_{lwa})$, as Eq.(\ref{equ:lambda}) shows.

\begin{scriptsize} 
\begin{equation}
  \label{equ:lambda}
  \begin{split}
  &\underset{\lambda}{\operatorname{\mathbf{argmin}}}
\left(\lambda ^2Cov(X_L)+(1-\lambda)^2Cov(X_G)\right)\\
\Rightarrow&\underset{\lambda}{\operatorname{\mathbf{argmin}}}
\left( (Cov(X_L)+Cov(X_G)) \lambda ^2 - 2Cov(X_G) \lambda + Cov(X_G)\right)\\
\Rightarrow& ((Cov(X_L)+Cov(X_G))\lambda = Cov(X_G) \\
  \end{split}
\end{equation}
\end{scriptsize}

Where $\lambda$ can be eliminated by combining Eq.(\ref{equ:lambdageometric}) and Eq.(\ref{equ:lambda}) to prove Eq.(\ref{equ:covXLiDAR}).

\subsection{LiDAR-SLAM Assisted GNSS RAIM Algorithm}
    In urban canyon environment, GNSS is seriously affected by the multipath and NLOS. 
    GNSS RAIM algorithm is normally adopted to detect the outliers of satellite observations. 
    But simultaneous presence of multiple outliers might underperform the existing GNSS RAIM algorithm. 
    In this subsection, LiDAR-SLAM assisted GNSS RAIM algorithm is proposed, the main flow of which is shown in 
    Algorithm \ref{alg:LiDAR-SLAM assisted GNSS RAIM algorithm}.
    Our RAIM Algorithm only uses pseudorange observations, so
    when the pseudorange of one satellite is detected as an outlier, 
    the phase observation of this observation is also considered an outlier.
    
\begin{algorithm}[t]
    \caption{LiDAR-SLAM assisted GNSS RAIM algorithm} 
    \label{alg:LiDAR-SLAM assisted GNSS RAIM algorithm}
    \hspace*{0.02in} {\bf Input:} 
    LiDAR-SLAM positioning result $X_L$, SPP result from ublox receiver $X_G$;\\
    \hspace*{0.02in} {\bf Output:} 
    GNSS Satellite Inliers;
    \begin{algorithmic}[1]
    \While{New GNSS Observation Comes}
        \State $X_F$ = SPP-LOAM($X_L$,$X_G$); //Fuse $X_L$ and $X_G$ by SPP-LOAM to get $X_F$ as fusion result;
        \State $dtr$ = $dtr$Est($X_F$, Pseudoranges); //Receiver clock error estimation;
        \State V=ResCal($X_F$, Pseudoranges, $dtr$); //Calculate pseudorange residuals by using Eq.(\ref{equ:SFPPP});
        \State [mean, std] = CalMeanStd(V); //For every GNSS satellite in this epoch, calculate the mean and standard deviation of its pseudorange residuals;
        \If{\big|V[j] - mean\big| $\leq$ $\alpha \cdot$std (//$\alpha$ is an input coefficient)}
        \State Set GNSS satellite j observations as inlier;
        \Else
        \State Set GNSS satellite j observations as outlier;
        \EndIf
  \EndWhile
    \State \Return GNSS Satellite Inliers;
    \end{algorithmic}
\end{algorithm}

\subsection{Pose Graph Model}
\begin{figure}[htp]
    \begin{center}
    \includegraphics[width=8.0cm,height=7.1129cm]{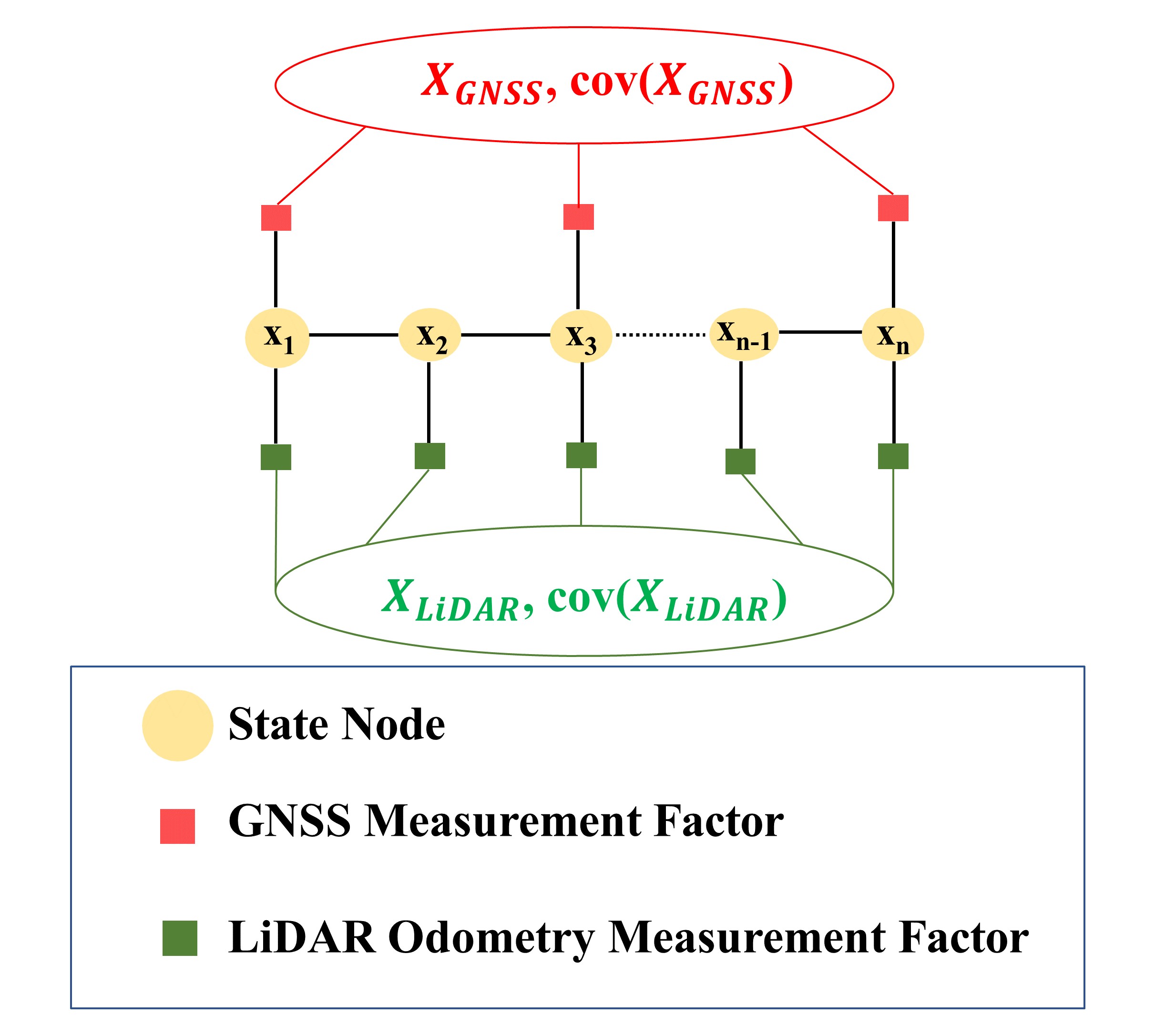}
    \end{center}
       \caption{Factor Graph of the Backend}
    \label{fig:Factor Graph of the Backend}
\end{figure}

In this subsection, we give the pose graph model in the form of factor graph (see Fig. \ref{fig:Factor Graph of the Backend}).
Then the specific optimization functions will be listed.

GNSS positioning results reflect the position of the GNSS antenna phase center 
which is represented as $\mathbf{P}_{G}^{\mathcal{N}}$ in Frame $\mathcal{N}$.
And the GNSS antenna phase center in Frame $\mathcal{L}$, also known as the extrinsic parameter, 
is represented as $\mathbf{P}_{G}^{\mathcal{L}}$. 
Other notations can be found in Table \ref{table:Notations}.
According to the coordinate transformation, we obtain the equation Eq.(\ref{equ:hengdengshi}).
\begin{equation}
\label{equ:hengdengshi}
    \mathbf{R}_{\mathcal{N}}^{\mathcal{M}}\mathbf{P}_{G}^{\mathcal{N}}+\mathbf{t}_{\mathcal{N}}^{\mathcal{M}}
        =\mathbf{R}_{\mathcal{L}}^{\mathcal{M}}\mathbf{P}_{G}^{\mathcal{L}}+\mathbf{t}_{\mathcal{L}}^{\mathcal{M}}
\end{equation}
According to Eq.(\ref{equ:hengdengshi}), the residual equation Eq.(\ref{equ:r1}) is obtained:
\begin{equation}
    \label{equ:r1}
    \begin{split}
    \begin{aligned}
    \mathbf{r}_{1}&=\mathbf{r}_{GNSS_{-}\text{LiDAR}}\\
    &=\mathbf{R}_{\mathcal{M}}^{\mathcal{N}}
		\left(\mathbf{R}_{\mathcal{L}}^{\mathcal{M}}\mathbf{P}_{G}^{\mathcal{L}}
		+\mathbf{t}_{\mathcal{L}}^{\mathcal{M}}-\mathbf{t}_{\mathcal{N}}^{\mathcal{M}}\right)-\mathbf{P}_{G}^{\mathcal{N}}
    \end{aligned}
    \end{split}
\end{equation}

And LiDAR-SLAM outputs a 6DoF pose with a certain frequency. This paper defines LiDAR-SLAM output as prior information.
The prior residual is listed in Eq.(\ref{equ:r2}) where $i$ and $j$ are two epochs when LiDAR observations are obtained.
\begin{equation}
    \label{equ:r2}
    \begin{array}{c}
    \mathbf{r}_{2}=\mathbf{r}_{LiDAR_{-}\text{prior}}\!=\!
    \end{array}
    \left[
    \begin{array}{c}
    \mathcal{R}^{\mathcal{L}}_{\mathcal{M},i}\left(\mathbf{t}_{\mathcal{L},j}^{\mathcal{M}}-\mathbf{t}_{\mathcal{L},i}^{\mathcal{M}}\right) \\
    \log \left(\mathcal{R}^{\mathcal{L}}_{\mathcal{M},i} \mathcal{R}_{\mathcal{L},j}^{\mathcal{M}}\right)^{\vee}
    \end{array}
    \right]
\end{equation}

The optimization function Eq.(\ref{equ:costfunction}) is obtained after combining Eq.(\ref{equ:r1}) and Eq.(\ref{equ:r2}).

\begin{equation}
    \label{equ:costfunction}
    \underset{\mathbf{R}_{\mathcal{N}}^{\mathcal{M}},\mathbf{t}_{\mathcal{N}}^{\mathcal{M}},
	\mathbf{R}_{\mathcal{L}}^{\mathcal{M}},\mathbf{t}_{\mathcal{L}}^{\mathcal{M}},\mathcal{P}_{G}^{\mathcal{L}}}{\operatorname{\mathbf{argmin}}}
	\left(\left\|\mathbf{r}_{1}\right\|_{\Sigma_{1}}^{2}+\left\|\mathbf{r}_{2}\right\|_{\Sigma_{2}}^{2}\right)
\end{equation}

The specific optimization functions for every observation epoch are listed in Eq.(\ref{observation_matrix}). 
Then we use Levenberg-Marquard algorithm to solve the cost function where a 15DoF variable needs to be optimized.
Its first six dimensions determine the coordinate transformation between Frame $\mathcal{N}$ and Frame $\mathcal{M}$.
The next six dimensions determine the pose of experimental vehicle.
The last three dimensions are the extrinsic parameters.

\begin{equation}
  \label{observation_matrix}
  \begin{split}
  \left[
  \begin{smallmatrix}
    \mathbf{r}_{1} \\
    \mathbf{r}_{2} \\
  \end{smallmatrix}
  \right] &=     
  \left[
  \begin{smallmatrix}
    \mathcal{R}_{\mathcal{M}}^{\mathcal{N}}\left(\mathcal{R}_{\mathcal{L}}^{\mathcal{M}} P_{G}^{\mathcal{L}}+\mathbf{t}_{\mathcal{L}}^{\mathcal{M}}-\mathbf{t}_{\mathcal{N}}^{\mathcal{M}}\right)^{\wedge} 
    & \boldsymbol{0_{3\times3}}\\
    -\mathcal{R}_{\mathcal{M}}^{\mathcal{N}} & \boldsymbol{0_{3\times3}} \\
    \mathcal{R}_{\mathcal{M}}^{\operatorname{\mathcal{N}}}\left(-\mathcal{R}_{\mathcal{L}}^{\mathcal{M}} P_{G}^{\mathcal{L}}\right)^{\wedge}
    & \mathcal{R}_{\mathcal{M}}^{\mathcal{L}} \\
    \mathcal{R}_{\mathcal{M}}^{\mathcal{N}} & \mathcal{R}_{\mathcal{M}}^{\mathcal{L}}\\
    \mathcal{R}_{\mathcal{M}}^{\mathcal{N}}\mathcal{R}_{\mathcal{L}}^{\mathcal{M}} &	\boldsymbol{0_{3\times3}}
  \end{smallmatrix}
  \right]^T       
  \left[
  \begin{smallmatrix}
    \delta \theta_{\mathcal{N}}^{\mathcal{M}} \\
    \delta \mathbf{t}_{\mathcal{N}}^{\mathcal{M}} \\
    \delta \theta_{\mathcal{L}}^{\mathcal{M}} \\
    \delta \mathbf{t}_{\mathcal{L}}^{\mathcal{M}} \\
  \delta \mathcal{P}_{G}^{\mathcal{L}}\\
  \end{smallmatrix}
  \right]     
  \end{split}
\end{equation}

All the data was collected and synchronized using the robot operation system (ROS).
But these two kinds of observations are not obtained at the same moment. 
In this case, we use linear extrapolation to obtain a virtual LiDAR observation which can be fused with GNSS.

\section{Experiment and analysis}
\subsection{Experiment Setup}
Our experiment used UrbanNav Dataset \cite{Weisong2020GNSS}, an open dataset collected by Hong Kong Polytechnic University.
In this dataset, a single frequency GNSS receiver, ublox M8T, was used to collect GNSS measurements
(only two systems, GPS and Beidou, were collected). 
The 3D LiDAR (Velodyne 32) was employed to collect 3D point clouds. 
In addition, the NovAtel SPAN-CPT, an RTK/INS integrated navigation system, 
was used to provide the ground truth of positioning.

\begin{figure}[htp]
  \centering
    \begin{center}
      \includegraphics[width=8.15cm,height=4.561cm]{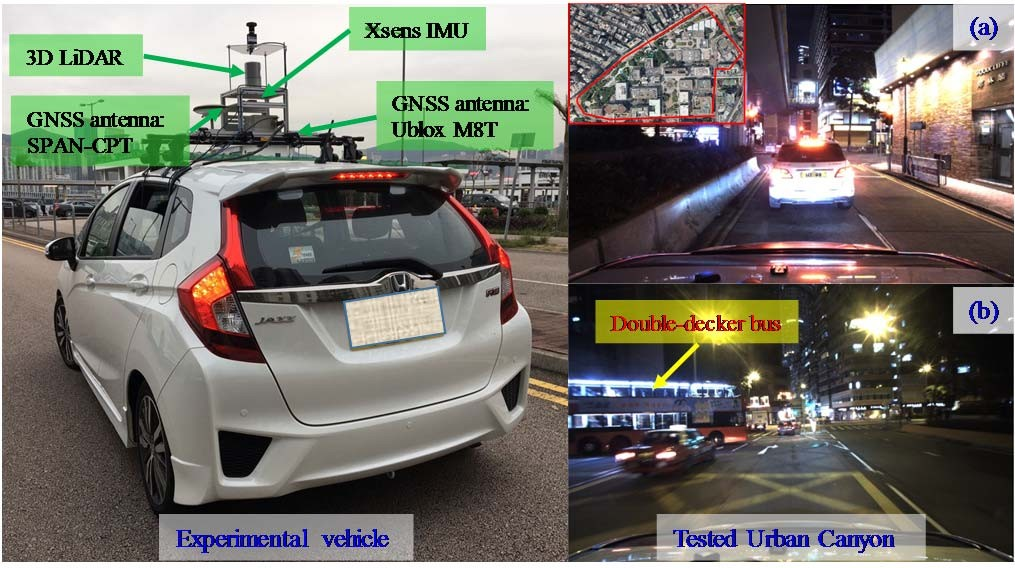}
    \end{center}
       \caption{Experimental Vehicle and the Tested 
       Scenario \cite{Weisong2020GNSS}}  
    \label{fig:Experimental_vehicle}
\end{figure}

Fig. \ref{fig:Experimental_vehicle}-(a) shows that the dataset was collected in an urban canyon environment, which is 
challenging for GNSS.
And Fig. \ref{fig:Experimental_vehicle}-(b) shows that there were many dynamic objects such as cars when the dataset was collected, 
which is challenging for LiDAR-SLAM.

\subsection{LiDAR-SLAM Positioning Covariance Estimation Results}
The estimated LiDAR-SLAM positioning covariance is in 
Frame $\mathcal{M}$ and we use Eq.(\ref{equ:covM2N}) to 
turn it into Frame $\mathcal{E}$.
\begin{equation}
    \label{equ:covM2N}
    \begin{cases}
        Cov(X_{L}^{\mathcal{N}}) = \mathbf{R}_{\mathcal{M}}^{\mathcal{N}}Cov(X_{L}^{\mathcal{M}})\mathbf{R}_{\mathcal{N}}^{\mathcal{M}}\\
        Cov(X_{L}^{\mathcal{E}}) = \mathbf{R}_{\mathcal{N}}^{\mathcal{E}}Cov(X_{L}^{\mathcal{N}})\mathbf{R}_{\mathcal{E}}^{\mathcal{N}}
    \end{cases}
\end{equation}

$\mathbf{R}_{\mathcal{M}}^{\mathcal{N}}$ is estimated from 
Eq.(\ref{equ:covM2N}). $B$ and $L$ in Eq.(\ref{equ:covM2N}) respectively represent 
the longitude and latitude of the current LiDAR-SLAM positioning results.
\begin{equation}
    \label{equ:covN2E}
    \mathbf{R}_{\mathcal{N}}^{\mathcal{E}} =
    \left[\begin{array}{ccc}
    -\sin B \cos L & -\sin B \sin L & \cos B \\
    -\sin L & \cos B & 0 \\
    \cos B \cos L & \cos B \sin L & \sin B
    \end{array}\right]
\end{equation}

\begin{figure}[htp]
  \begin{center}
  \includegraphics[width=7.38cm,height=12.69cm]{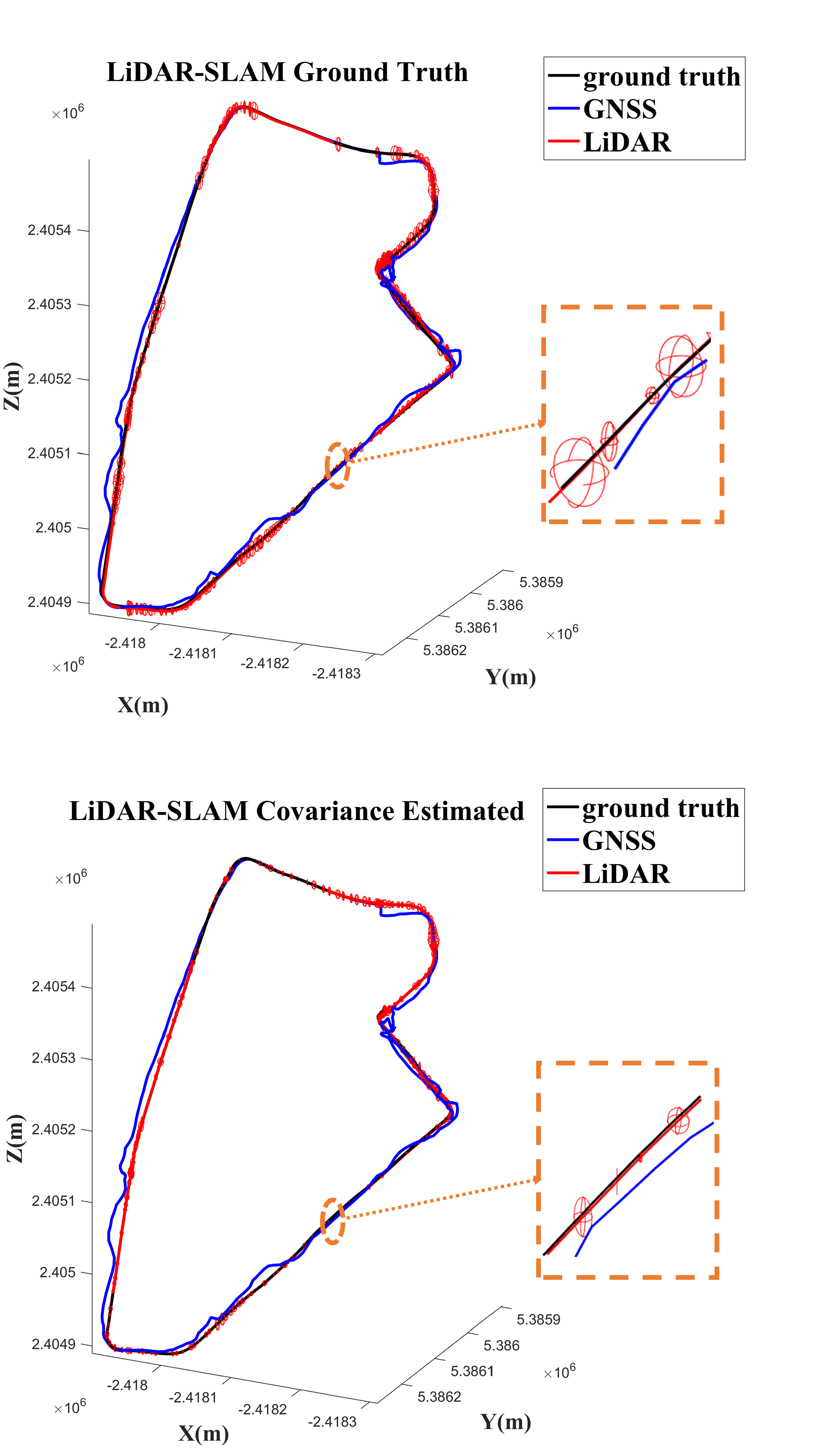}
  \end{center}
     \caption{LiDAR-SLAM Covariance Results}
  \label{fig:LidarSLAMcov}
\end{figure}

The LiDAR-SLAM positioning covariance estimation results are shown 
in Fig. \ref{fig:LidarSLAMcov}, 
in which the black is the ground truth trajectory, 
the blue is the GNSS trajectory, and the red is the LiDAR-SLAM trajectory.
The red ellipse in the upper part of Fig. \ref{fig:LidarSLAMcov} 
represents the LiDAR-SLAM covariance ground truth calculated by 
Eq.(\ref{equ:covXLiDAR}).
The red ellipse in the bottom part of Fig. \ref{fig:LidarSLAMcov} 
shows the estimated LiDAR-SLAM covariance calculated by 
Eq.(\ref{equ:lambdat}).

In Table \ref{table:LidarSLAMcovtable}, six groups of values, each composed of the mean and standard deviation, 
are displayed to represent the difference between the ground truth and estimated LiDAR-SLAM covariance. 
The means and the standard deviations fall within the range of 0.36m$^2$ and 0.54m$^2$ respectively, 
which are relatively small differences comparing to the positioning error.

\begin{table}[htp]
    \centering
    \caption{The Difference Between the Ground Truth and the Estimated Value of LiDAR-SLAM Covariance}
    \setlength{\tabcolsep}{0.8mm}{
    \begin{tabular*}{239pt}{cccccccc}
    \hline
    \hline
            &\makecell[c]{$\Delta$CovX}&\makecell[c]{$\Delta$CovY}&\makecell[c]{$\Delta$CovZ}
            &\makecell[c]{$\Delta$CovXY}&\makecell[c]{$\Delta$CovYZ}&\makecell[c]{$\Delta$CovZX}\\
    \hline
    Mean($m^2$)     &0.2365  & 0.3519  & 0.2125 & 0.2831 & 0.2716 & 0.2125 \\
    Std($m^2$)      &0.4785  & 0.5320  & 0.3575 & 0.4824 & 0.4286 & 0.3575 \\
    \hline
    \hline
    
    \end{tabular*}}
    \label{table:LidarSLAMcovtable}
\end{table}

\subsection{LiDAR-SLAM Assisted GNSS RAIM Performance}
In this subsection, we will compare the SF-PPP performance 
before and after using 
RAIM (Algorithm \ref{alg:LiDAR-SLAM assisted GNSS RAIM algorithm}).

\begin{figure}[htp]
  \begin{center}  
  \includegraphics[width=8.30cm,height=7.023cm]{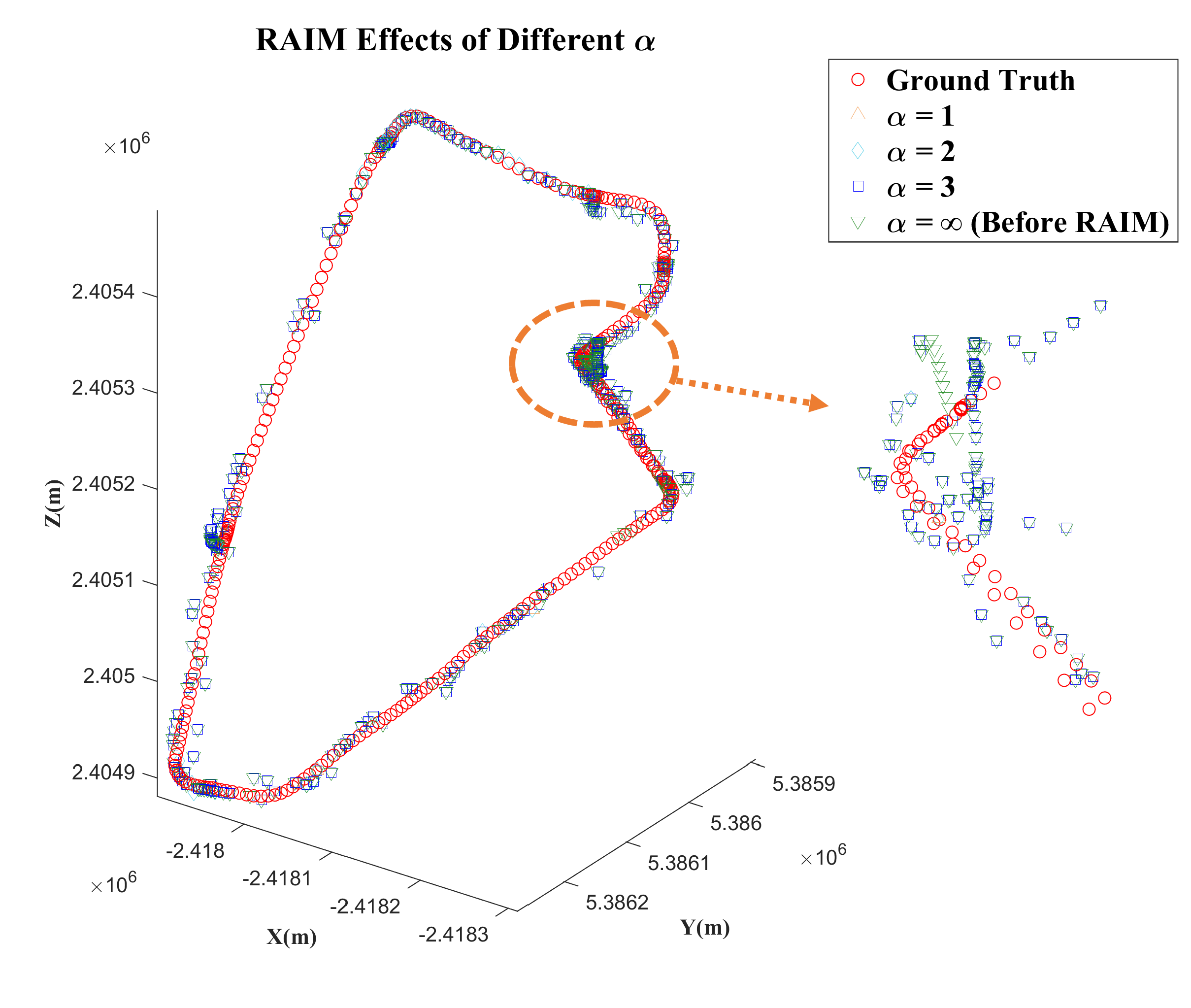}
  \end{center}
     \caption{RAIM Effects of Different $\alpha$}
  \label{fig:SF-PPP Result Before and After RAIM}
\end{figure}

When the coefficient $\alpha$ in Algorithm \ref{alg:LiDAR-SLAM assisted GNSS RAIM algorithm} is larger, the effect of RAIM gets smaller. 
Before using RAIM, the SF-PPP in urban canyon environment has low reliability, 
as the green points shown in Fig. \ref{fig:SF-PPP Result Before and After RAIM}. 
When the coefficient $\alpha$ gets smaller, the remaining number of positioning results decreases. 
Different $\alpha$ values have been tried, and finally, $\alpha=2$ is chosen considering the availability and reliability.

\begin{figure}[htp]
    \begin{center}
    \includegraphics[width=7.0cm,height=5.252cm]{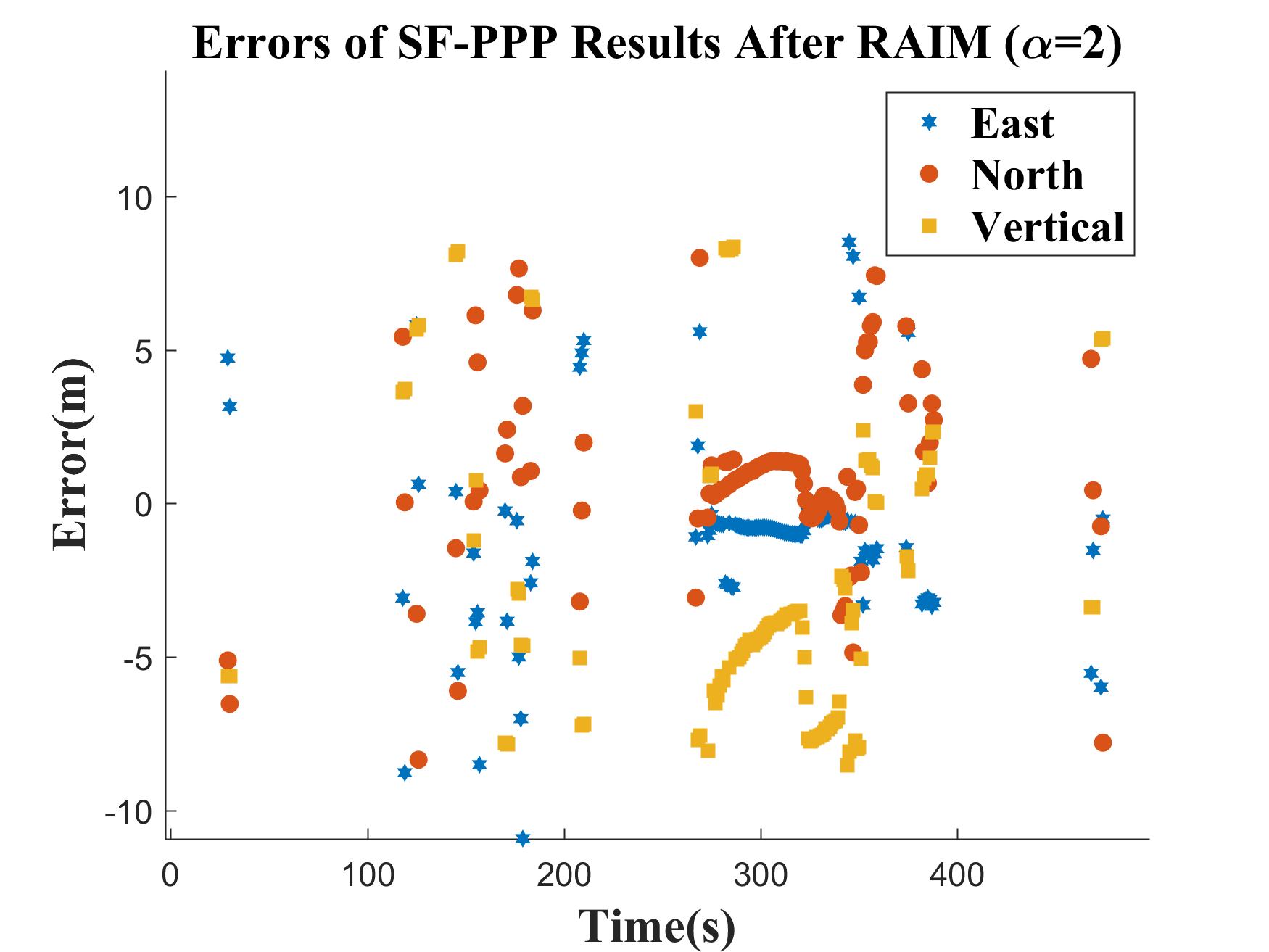}
    \end{center}
       \caption{Errors of SF-PPP Results After RAIM}
    \label{fig:SF-PPP Result After RAIM}
\end{figure}

The error of PPP (when $\alpha=2$) is plotted in Fig. \ref{fig:SF-PPP Result After RAIM}.
We find that the availability and accuracy of PPP are much higher between 270 to 390 seconds 
according to Fig. \ref{fig:SF-PPP Result After RAIM}.
This is due to better GNSS observation conditions where there are fewer buildings.

\subsection{P$^3$-LOAM Performance}
In order to prove the validity of P$^3$-LOAM, several different positioning schemes were compared on UrbanNav. 
Specific statistical results are in Table \ref{table:GLOAM Result}. 
The column of `LeGO-LOAM' represents LiDAR-SLAM's results in local coordinates. 
`LeGO-LOAM'\footnote{https://github.com/RobustFieldAutonomyLab/LeGO-LOAM} is an open-source software widely used as a benchmark. 
The column of `Hsu et al.' represents the GNSS and LiDAR-SLAM fusion results reported in \cite{Weisong2020GNSS}. 
The column of `SPP' represents the GNSS results processed by SPP. 
The column of `PPP' represents the GNSS results processed by SF-PPP with LiDAR-SLAM assisted RAIM algorithm. 
The results of `PPP' are shown in Fig. \ref{fig:SF-PPP Result After RAIM}. 
`SPP' and `PPP' are both from RTKLIB\footnote{https://github.com/tomojitakasu/RTKLIB}, 
a piece of software often used for GNSS research. 
The column of `SPP-LOAM' represents the SPP and LiDAR-SLAM fusion results. 
The column of `P$^3$-LOAM' represents the SF-PPP and LiDAR-SLAM fusion results.

The first four rows of Table \ref{table:GLOAM Result} represent error indicators.
We use $X$ to represent the ground truth and $\tilde{X}$ to represent the estimated results.
The `MAE' is short for `Mean Absolute Error'.
It can be calculated by Eq.(\ref{equ:MAE}).
\begin{equation}
  \label{equ:MAE}
  \operatorname{MAE}(\mathrm{\tilde{X}}, \mathrm{X})=\frac{1}{\mathrm{m}} \sum_{i=1}^{\mathrm{m}}
  \left|\mathrm{\tilde{X}}^{(\mathrm{i})}-\mathrm{X}^{(\mathrm{i})}\right|
\end{equation}
The `RMSE' is short for `Root Mean Square Error'.
It can be calculated by Eq.(\ref{equ:RMSE}).
\begin{equation}
  \label{equ:RMSE}
  \operatorname{RMSE}(\mathrm{\tilde{X}}, \mathrm{X})=\frac{1}{\mathrm{m}} \sum_{i=1}^{\mathrm{m}}\left(
    \mathrm{\tilde{X}}^{(\mathrm{i})}-\mathrm{X}^{(\mathrm{i})}\right)^2
\end{equation}
The `Max' represents the max error.
The `Std' is short for `Standard Deviation' and represents the standard deviation of error which can be 
calculated by Eq.(\ref{equ:Std}).
\begin{equation}
  \label{equ:Std}
  \operatorname{Std}=\sqrt{\frac{1}{m} \sum_{i=1}^{m}\left(\left|\mathrm{\tilde{X}}^{(\mathrm{i})}-\mathrm{X}^{(\mathrm{i})}\right|-\operatorname{MAE}\right)^{2}}
\end{equation}

The `Scale' row in Table \ref{table:GLOAM Result} represents the ability of algorithms to calculate global coordinates.

The `Availability' is calculated by dividing positioning result numbers by observation numbers.
\begin{table}[htp]
    \caption{GNSS and LiDAR-SLAM Fusion Results}
    \begin{threeparttable}  
    \centering
    \setlength{\tabcolsep}{0.8mm}{
    \begin{tabular*}{252pt}{cccccccc}
    \hline
    \hline
            &\makecell[c]{LeGO-\\LOAM \cite{legoloam2018}}&
            \makecell[c]{Hsu et al.\\ \cite{Weisong2020GNSS}}&\makecell[c]{SPP}  
            &\makecell[c]{PPP}&\makecell[c]{SPP-LOAM}&\makecell[c]{P$^3$-LOAM}\\
    \hline
    MAE(m)&4.28  & 4.60  & 21.18 & 5.92 & 20.71 & \textbf{3.44} \\
    RMSE(m)      &4.50  & -$^*$    & 27.56 & 6.17 & 23.37 & \textbf{3.72} \\
    Max(m)      &8.27  & 16.46 & 88.04 & 8.95 & 36.97 & \textbf{7.44} \\
    Std(m)      &\textbf{1.40}  & 3.33  & 17.64 & 1.73 & 7.55 &1.43 \\
    Scale      &Local    &Global  & Global & Global &Global  & \textbf{Global}\\
    Availability &$100\%$ &$100\%$ &$100\%$ &$21.4\%$ &$100\%$ &$\textbf{100\%}$ \\
    \hline
    \hline
    \end{tabular*}}
    \begin{tablenotes}
    \item[*] The `-' means the result was not reported.
    \end{tablenotes}
    \end{threeparttable} 
    \label{table:GLOAM Result}
\end{table}

The results in Table \ref{table:GLOAM Result} show that P$^3$-LOAM performs best in almost every indicator except in the `Std' where
LeGO-LOAM and P$^3$-LOAM have the same level of performance. We analyze that the errors of LeGO-LOAM are larger but more concentrated since 
SLAM has the property of error accumulation.

\begin{figure*}[htp]
    \begin{center}
    \includegraphics[width=15.00cm,height=8.45775cm]{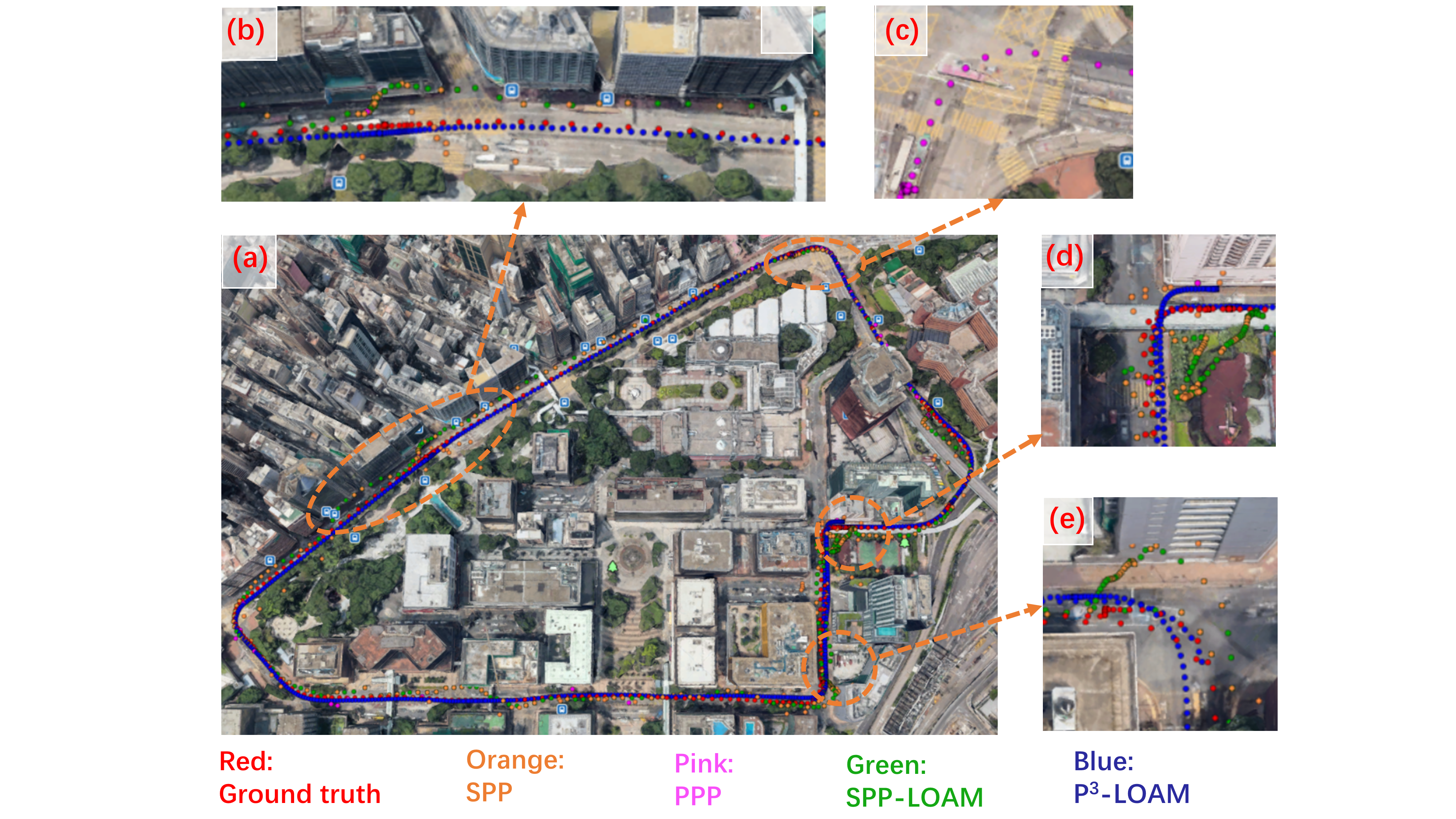}
    \end{center}
       \caption{All the Positioning Results on Google Earth}
    \label{fig:RESULT}
\end{figure*}

\begin{figure}[htp]
  \begin{center}
  \includegraphics[width=8cm,height=4.7758cm]{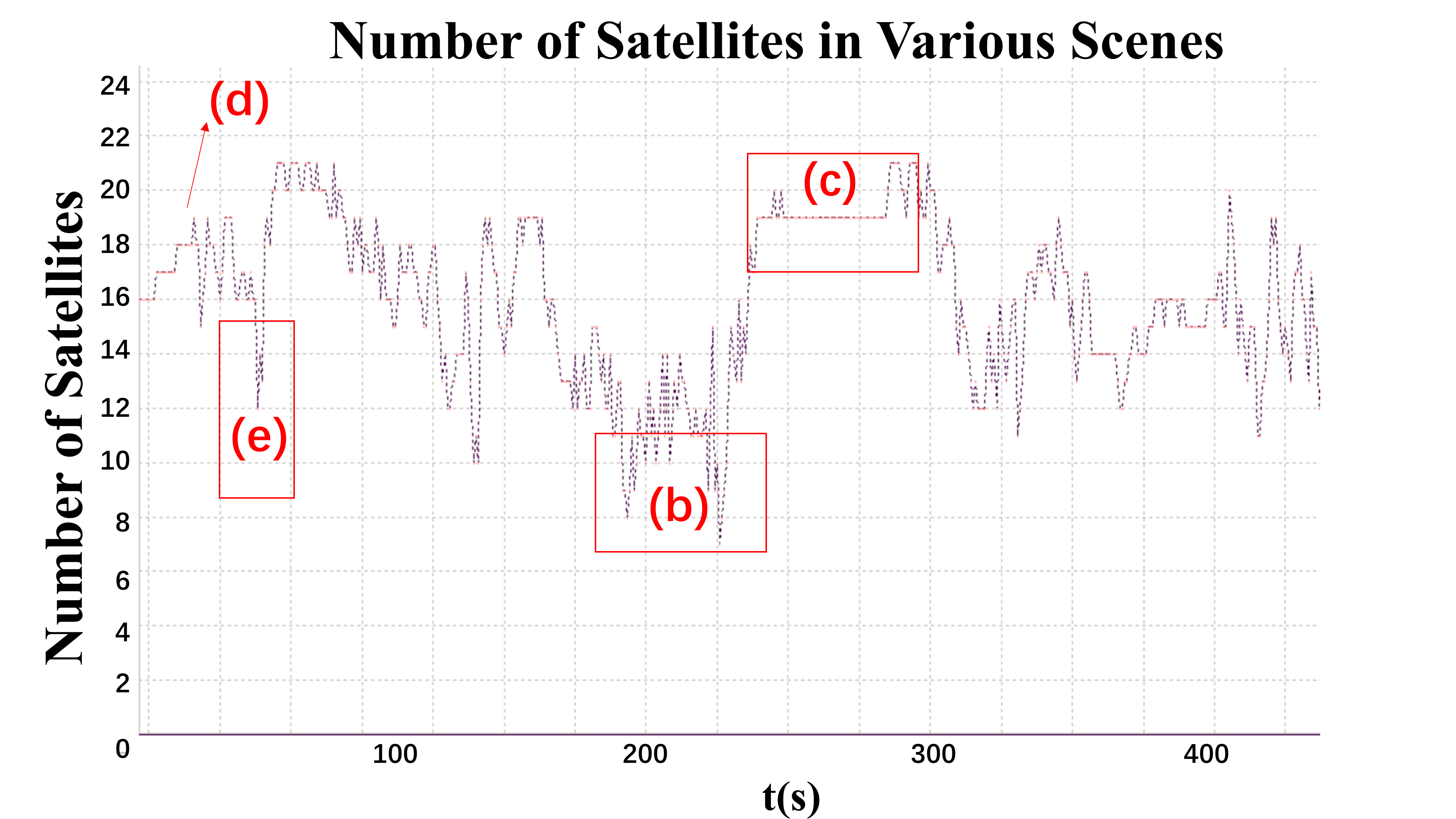}
  \end{center}
     \caption{Number of Satellites}
  \label{fig:NoS}
\end{figure}

Fig. \ref{fig:RESULT}-(a) shows all the results plotted on Google Earth. 
It is evident that the scene is challenging for GNSS. Especially in Fig. \ref{fig:RESULT}-(b), (d), (e), 
where the tall buildings exert negative effects on satellite signals, so SPP and SPP-LOAM cannot
perform well here whereas P$^3$-LOAM still provides reliable results.
In contrast, the scene in Fig. \ref{fig:RESULT}-(c) is friendly to GNSS, so all the results here are close to the ground truth 
and the PPP results here are more available than in any other parts. 
Fig. \ref{fig:NoS} displays the number of available satellites to account for the impact of the scene on GNSS.
We can see that Fig. \ref{fig:NoS}-(b) and Fig. \ref{fig:NoS}-(e) have fewer satellites observed than the others, 
leading to the unsatisfactory GNSS results in these parts.
Fig. \ref{fig:NoS}-(c) and Fig. \ref{fig:NoS}-(d) both have adequate satellites, but the performance in Fig. \ref{fig:NoS}-(d) 
is severely degraded by the multipath effect or NLOS caused by the tall buildings.

\begin{figure}[htp]
  \begin{center}
  \includegraphics[width=8.0cm,height=6.0cm]{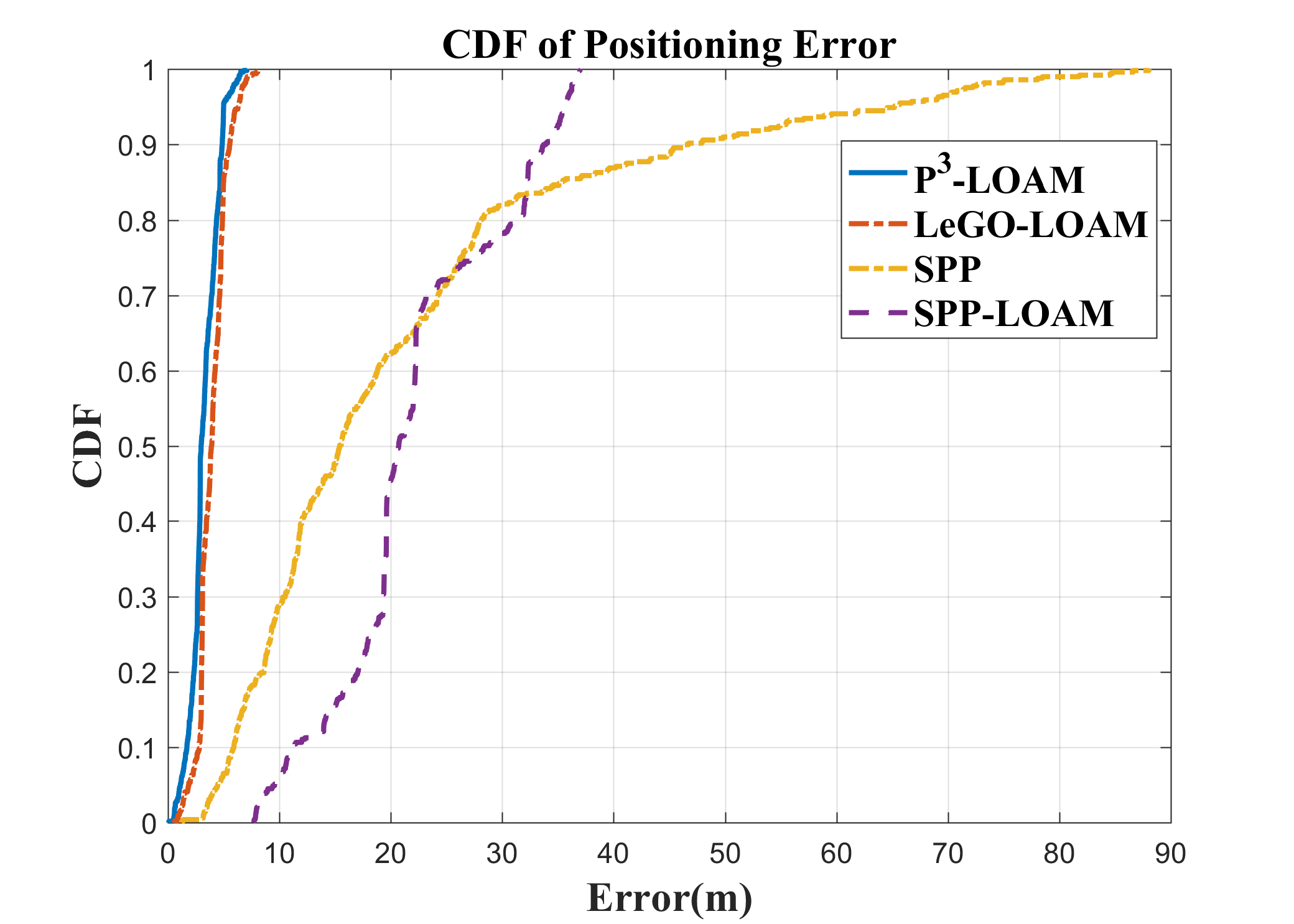}
  \end{center}
     \caption{CDF of Positioning Error}
  \label{fig:CDF}
\end{figure}

\begin{figure}[htp]
  \begin{center}
  \includegraphics[width=8.1cm,height=5.9372cm]{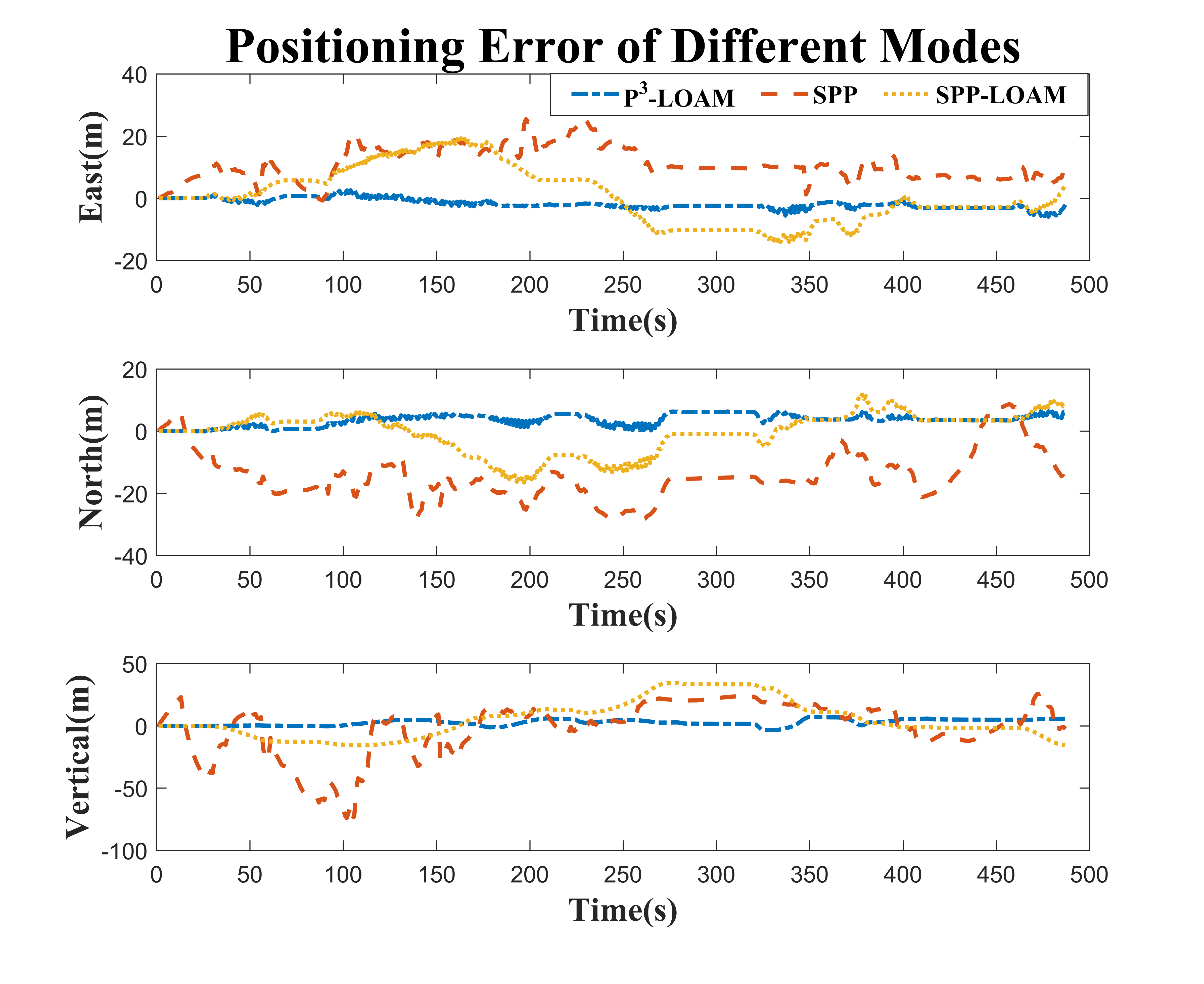}
  \end{center}
     \caption{Positioning Error of Different Modes}
  \label{fig:ENU}
\end{figure}

The availability of `LeGO-LOAM', `SPP', `SPP-LOAM' and `P$^3$-LOAM' is all 100\% which indicates the continuity of algorithms. 
To compare the availability of these four methods, we further analyze the Cumulative Distribution Function(CDF) of
their errors in Fig. \ref{fig:CDF}. If we define the availability as the error under an exact level, such as 5 meters, 
the availability of `SPP' and `SPP-LOAM' will be much lower than `P$^3$-LOAM' and `LeGO-LOAM'. 
The errors in the east, north and vertical components are also plotted in Fig. \ref{fig:ENU}.
It is obvious that `P$^3$-LOAM' performs best when considering accuracy, availability, and the ability to obtain global coordinates.

At last, we count the runtime of the proposed method. 
The computation loads of the five modules in Fig. \ref{fig:SystemOverview} are listed in Table \ref{tab:Runtime}. 
It is noted that the CPU of test computer is i7-7700. 
The LiDAR-SLAM module is derived from `LeGO-LOAM', and other modules bring the additional time consumption about 12\% of LiDAR-SLAM. 
It means that LiDAR-SLAM module is still the dominant computational load of proposed method 
which is nearly as much as LeGO-LOAM's computation. 

\begin{table}[htbp]
  \centering
  \caption{Runtime of Modules}
    \begin{tabular}{cc}
    \hline
    \hline   
    Modules & Runtime Per Second(ms) \\ 
    \hline 
    LiDAR-SLAM & 340.7 \\
    LiDAR-SLAM Covariance Estimation & 2.5 \\
    PPP Algorithm & 17.3 \\
    LiDAR-SLAM Assisted GNSS RAIM & 10.5 \\
    15DoF Variable Optimization & 10.2 \\
    \hline 
    \hline
    \end{tabular}%
  \label{tab:Runtime}%
\end{table}%

\section{Conclusion}
In order to obtain robust and reliable positioning results in urban canyon environment, we couple LiDAR and PPP in this paper.
For better fusion, we have done the following work:
Firstly, we derive an SVD Jacobian based error propagation model to 
estimate the covariance of LiDAR-SLAM, and the result is evaluated 
using GNSS covariance and ground truth of trajectory.
Secondly, we propose a LiDAR-SLAM assisted GNSS RAIM
algorithm, achieving reliable GNSS positioning results in urban canyon environment.
Thirdly, P$^3$-LOAM is proposed, 
combining PPP and LiDAR-SLAM with accurate covariance estimation.
The performance of the proposed methods is assessed with a series of experiments: 
1) LiDAR-SLAM positioning covariance is evaluated and it is proved to be consistent with that of GNSS.  
2) SF-PPP error is gratifyingly controlled using LiDAR-SLAM assisted GNSS RAIM algorithm.
3) We fuse PPP and LiDAR-SLAM to be P$^3$-LOAM, 
producing better positioning results than any other methods.

However, the availability of PPP in this paper is not satisfactory. 
In future research, we will work on a tighter coupled LiDAR-SLAM/PPP navigation system 
so as to improve the availability of PPP in urban canyon environment.
In addition, dual-antenna GNSS, or even more antennas, will be set in the vehicle to obtain 
a global posture for better fusion with LiDAR-SLAM.
\section*{Acknowledgment}
This work was supported by the National Nature Science Foundation of China (NSFC) under Grant 61873163, and
Equipment Pre-ResearchField Foundation under Grant 61405180205 and Grant 61405180104.

\ifCLASSOPTIONcaptionsoff
  \newpage
\fi

{\small
\normalem
\bibliographystyle{ieeetr}
\bibliography{egbib}
}
\begin{IEEEbiography}[{\includegraphics[width=1in,height=1.25in,clip,keepaspectratio]{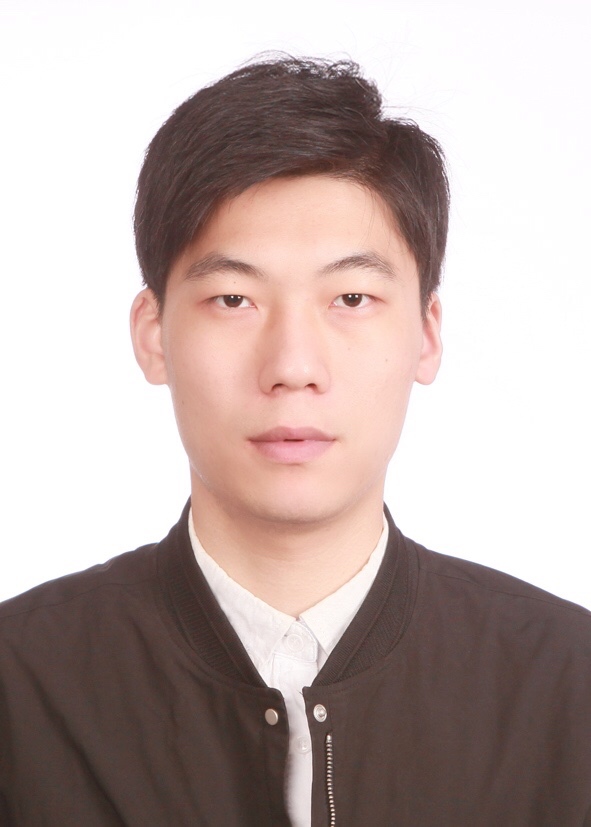}}]{Tao Li}
  received the B.S. degree in navigation engineering from Wuhan University, Wuhan, China, in 2018. 
  He is currently pursuing the Ph.D. degree with Shanghai Jiao Tong University, Shanghai, China. 
  His current research interests include Visual-SLAM, 
  LiDAR-SLAM, global navigation satellite systems(GNSS), 
  inertial navigation systems (INS), and information fusion.
\end{IEEEbiography}
\begin{IEEEbiography}[{\includegraphics[width=1in,height=1.25in,clip,keepaspectratio]{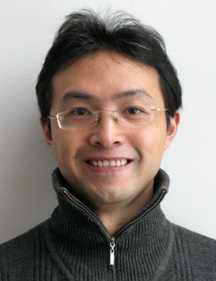}}]{Ling Pei}
received the Ph.D. degree from Southeast University, Nanjing, China, in 2007.From 2007 to 2013, 
he was a Specialist Research Scientist with the Finnish Geospatial Research Institute. 
He is currently an Associate Professor with the School of Electronic Information and Electrical Engineering, Shanghai Jiao Tong University. 
He has authored or co-authored over 90 scientific papers. He is also an inventor of 24 patents and pending patents. 
His main research is in the areas of indoor/outdoor seamless positioning, ubiquitous computing, wireless positioning, 
Bio-inspired navigation, context-aware applications, location-based services, 
and navigation of unmanned systems. Dr. Pei was a recipient of the Shanghai Pujiang Talent in 2014.
\end{IEEEbiography}
\begin{IEEEbiography}[{\includegraphics[width=1in,height=1.25in,clip,keepaspectratio]{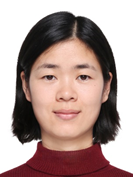}}]{Yan Xiang}
  is an assistant professor at Shanghai Jiao Tong University. 
  She obtained her Ph.D. in the Department of Geomatics Engineering at University of Calgary. 
  Her current research interests include inter-frequency and inter-system code and phase bias determination, 
  carrier-phase based ionospheric modeling, and ionospheric augmentation for high precision positioning.
  \end{IEEEbiography}
\begin{IEEEbiography}[{\includegraphics[width=1in,height=1.25in,clip,keepaspectratio]{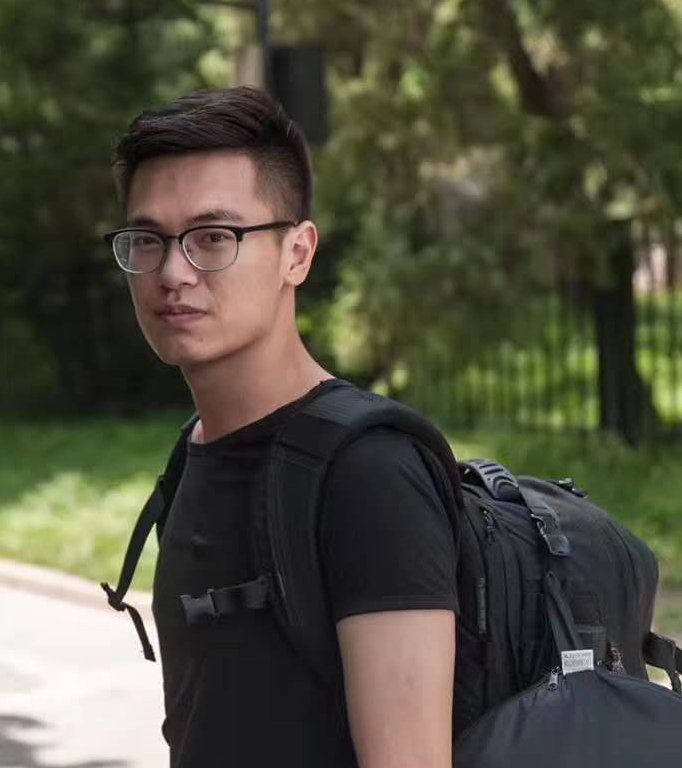}}]{Qi Wu}
  received the B.S. degree in Chongqing University of Posts and Telecommunications, Chongqing, China, 
  in 2018.He received the M.S. degree in Beijing University of Posts and Telecommunications, 
  Beijing, China. He is currently working toward the
  Ph.D degree in Shanghai Jiao Tong University. 
  His main research interests include visual-SLAM ,LiDAR-SLAM, Multi-sensor fusion.
\end{IEEEbiography}
\begin{IEEEbiography}[{\includegraphics[width=1in,height=1.25in,clip,keepaspectratio]{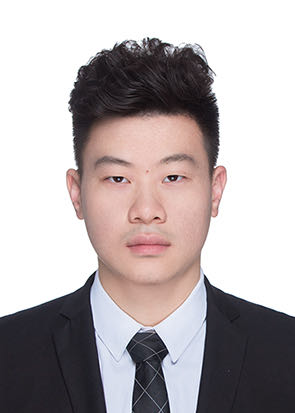}}]{Songpengcheng Xia}
  received the B.S. degree in navigation engineering from Wuhan University, Wuhan, China, in 2019. 
  He is currently pursuing the Ph.D. degree with Shanghai Jiao Tong University, Shanghai, China. \\
  His current research interests include machine learning, inertial navigation, 
  multi-sensor fusion and mixed-reality simulation technology.
\end{IEEEbiography}
\begin{IEEEbiography}[{\includegraphics[width=1in,height=1.25in,clip,keepaspectratio]{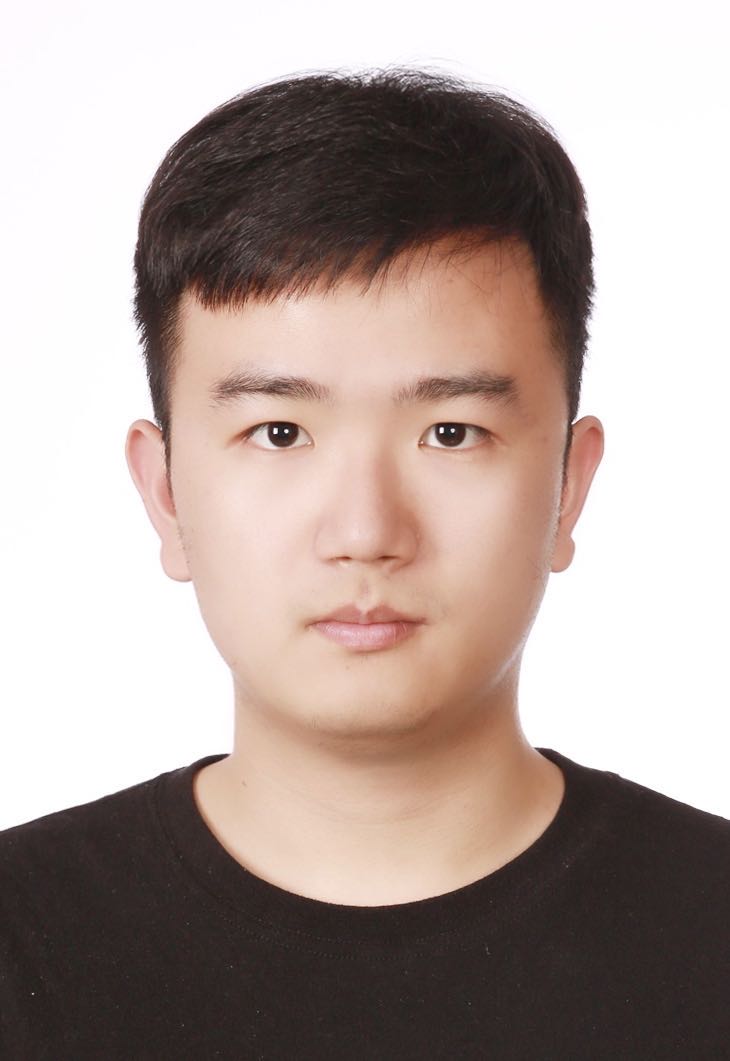}}]{Lihao Tao}
  received the B.S. degree in electronic engineering from Beihang University, Beijing, China, in 2018. 
  He is currently pursuing the M.S. degree with Shanghai Jiao Tong University, Shanghai, China. \\
  His current research interests include multiple sensors extrinsic calibration(IMU, Camera and LiDAR) and 
  LiDAR-SLAM.
\end{IEEEbiography}

\begin{IEEEbiography}[{\includegraphics[width=1in,height=1.25in,clip,keepaspectratio]{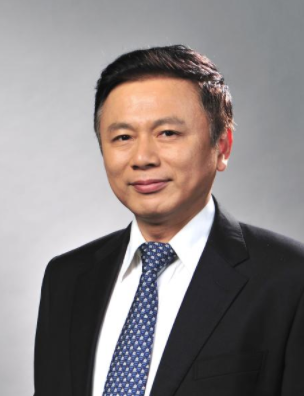}}]{Wenxian Yu}
  received the B.S., M.S., and Ph.D. degrees from the National University of Defense Technology, Changsha, China, 
  in 1985, 1988, and 1993, respectively. From 1996 to 2008, he was a Professor with the College of Electronic Science 
  and Engineering, National University of Defense Technology, where he was also the Deputy Head of the 
  College and an Assistant Director of the National Key Laboratory of Automatic Target Recognition. 
  From 2009 to 2011, he was the Executive Dean of the School of Electronic, Information, and Electrical Engineering, 
  Shanghai Jiao Tong University, Shanghai, China. He is currently a Yangtze River Scholar Distinguished Professor 
  and the Head of the research part in the School of Electronic, Information, and Electrical Engineering, Shanghai Jiao Tong University. 
  His research interests include remote sensing information processing, automatic target recognition, multi-sensor data fusion, etc.
\end{IEEEbiography}

\end{document}